\documentclass{article}

\usepackage{microtype}
\usepackage{graphicx}
\usepackage{subfigure}
\usepackage{booktabs}
\usepackage{enumitem}
\usepackage[title]{appendix}

\usepackage{listings}
\input{shortcuts.tex}
\usepackage{amsmath,amsthm,amsfonts,amssymb,amscd}
\usepackage{enumerate}
\usepackage{fancyhdr}
\usepackage{mathrsfs}
\usepackage{xcolor}
\usepackage{graphicx}
\usepackage{listings}
\usepackage{amsmath}
\usepackage{hyperref}
\usepackage{mathtools}
\usepackage{balance}

\newcommand{\PD}[2]{\frac{\partial{#1}}{\partial{#2}}}

\usepackage{hyperref}

\usepackage[accepted]{icml2021}

\icmltitlerunning{Efficient Differentiable Simulation of Articulated Bodies}

\begin{document}

\twocolumn[
\icmltitle{Efficient Differentiable Simulation of Articulated Bodies}

\icmlsetsymbol{equal}{*}

\begin{icmlauthorlist}
\icmlauthor{Yi-Ling Qiao}{equal,umd,intel}
\icmlauthor{Junbang Liang}{equal,umd}
\icmlauthor{Vladlen Koltun}{intel}
\icmlauthor{Ming C. Lin}{umd}
\end{icmlauthorlist}

\icmlaffiliation{umd}{University of Maryland, College Park}
\icmlaffiliation{intel}{Intel Labs}

\icmlcorrespondingauthor{Yi-Ling Qiao, Junbang Liang, and Ming C. Lin}{\{yilingq,liangjb,lin\}@cs.umd.edu}
\icmlcorrespondingauthor{Vladlen Koltun}{vladlen.koltun@intel.com}

\icmlkeywords{differentiable physics}

\vskip 0.3in
]
\printAffiliationsAndNotice{\icmlEqualContribution}
\begin{abstract}
We present a method for efficient differentiable simulation of articulated bodies. This enables integration of articulated body dynamics into deep learning frameworks, and gradient-based optimization of neural networks that operate on articulated bodies. We derive the gradients of the forward dynamics using spatial algebra and the adjoint method. Our approach is an order of magnitude faster than autodiff tools. By only saving the initial states throughout the simulation process, our method reduces memory requirements by two orders of magnitude. We demonstrate the utility of efficient differentiable dynamics for articulated bodies in a variety of applications. We show that reinforcement learning with articulated systems can be accelerated using gradients provided by our method. In applications to control and inverse problems, gradient-based optimization enabled by our work accelerates convergence by more than an order of magnitude.
\end{abstract}

\section{Introduction}
\label{sec:introduction}
Differentiable physics enables efficient gradient-based optimization with dynamical systems. It has achieved promising results in both simulated~\cite{Hu2019:ICRA,Qiao2020Scalable} and real environments~\cite{bern2019trajectory,Learning2020Song}. Our goal is to make articulated body simulation efficiently differentiable. We aim to maximize efficiency in both computation and memory use, in order to support fast gradient-based optimization of differentiable systems that interact with articulated bodies in physical environments.

Articulated bodies play a central role in robotics, computer graphics, and embodied AI. Many control systems are optimized via experiences collected in simulation~\cite{todorov2012mujoco,Coumans2015bullet,lee2018dart,drake}.
However, they do not have access to analytic derivatives of the articulated dynamics. Therefore, nearly all gradient-based approaches have to design strategies to compute the gradients indirectly when dealing with articulated bodies.

The straightforward way to differentiate the simulation is to use existing automatic differentiation tools~\cite{griewank2008evaluating,tensorflow2016,Paszke2019pytorch}. However, autodiff tools consume prohibitive amounts of memory when there are many simulation steps. Autodiff tracks every operation and stores the intermediate results in order to perform backpropagation. In articulated body simulation, the iterative contact solver and the dynamics algorithm~\cite{Featherstone2007RigidBD} yield exceedingly long computational graphs. In our experiments, differentiable simulators built with autodiff tools run out of memory after 5,000 simulation steps~-- just a few seconds of experience. As a result, learning is constrained to short experiences or forced to used large time steps, thus curtailing the scope of behaviors that can be learned or undermining the simulation's stability.

Furthermore, the overhead of creating and storing the auxiliary variables for autodiff also slows down the forward simulation. Although autodiff tools like DiffTaichi~\cite{Hu2019:ICLR} and JAX~\cite{jax2018github} can accelerate the simulation of fluids and deformable solids by vectorization, it is difficult to achieve the same speedup in articulated body simulation because the articulated dynamics algorithm is highly serialized, unlike inherently parallel computation on grids and particles.

In this paper, we design a differentiable articulated body simulation method that runs 10x faster with 1\% of the memory consumption compared to differentiation based on autodiff tools.
In order to minimize the overhead of differentiation, we derive the gradients of articulated body simulation using the adjoint method~\cite{giles2000introduction}. 
\ylq{The adjoint method has been applied to fluids~\cite{mcnamara2004fluid} and multi-body systems~\cite{Geilinger2020add}, but these applications do not support physically correct differentiable simulation of articulated bodies.}
Our derivation of the operator adjoints uses spatial algebra~\cite{Featherstone2020spatial}. Our method needs almost no additional computation during the forward simulation and is an order of magnitude faster than autodiff tools.

We further reduce memory requirements by adapting ideas from checkpointing~\cite{GriewankWalther2000,chen2016training} to the differentiable simulation setting. In the forward pass, we store the initial simulation state for each time step. During backpropagation, we recreate intermediate variables by reproducing simulation from the stored state. The overall runtime remains fast, while memory consumption is reduced by two orders of magnitude.


As an application of differentiable dynamics for articulated systems, we show that reinforcement learning (RL) can benefit from the knowledge of gradients in two ways.
First, it can make use of the gradients computed by the simulation to generate extra samples using first-order approximation.
Second, during the policy learning phase, differentiable physics enables us to perform a one-step rollout of the objective value function so that the policy updates can be more accurate.
Both schemes effectively improve the convergence speed and the attained reward.
We also demonstrate applications of efficient differentiable articulated dynamics to inverse problems, such as motion control and parameter estimation. Gradient-based optimization enabled by our method accelerates convergence in these settings by more than an order of magnitude.  Code is available on our project page: \url{https://github.com/YilingQiao/diffarticulated}


The contributions of this work are as follows:
\begin{itemize}[noitemsep,topsep=0pt]
    \item We derive the adjoint formulations for the entire articulated body simulation workflow, enabling a 10x acceleration over autodiff tools.
    \item We adapt the checkpointing method to the structure of articulated body simulation to reduce memory consumption by 100x, making stable collection of extended experiences feasible.
    \item We introduce two general schemes for accelerating reinforcement learning using differentiable physics.
    \item We demonstrate the utility of differentiable simulation of articulated bodies in motion control and parameter estimation, enhancing performance in these scenarios by more than an order of magnitude.
\end{itemize}

\section{Related Work}
\label{sec:related}

Differentiable programming has been applied to rendering~\cite{li2018diff,david2019mitsuba,Laine2020}, image processing~\cite{Li:2020:DVG,Li:2018:DPI}, SLAM~\cite{gradSLAM}, and design~\cite{du2020stokes}. Making complex systems differentiable enables learning and optimization using gradient-based methods.
Our literature review focuses on differentiable physics, the adjoint method, and neural approximations of physical systems.


\mypara{Differentiable physics.}
Differentiable physics provides gradients for learning, control, and inverse problems that involve physical systems. \citet{Degrave2019} advocate using differentiable physics to solve control problems in robotics. \citet{Belbute2018} obtain gradients of 2D rigid body dynamics. \citet{Liang2019} use automatic differentiation tools to obtain gradients for cloth simulation. \citet{Qiao2020Scalable} develop a more comprehensive differentiable physics engine for rigid bodies and cloth based on mesh representations. \citet{ingraham2019learning} present differentiable simulation of protein dynamics. For volumetric data, ChainQueen~\cite{Hu2019:ICRA} computes gradients of the MPM simulation pipeline. \citet{bern2019trajectory} use FEM to model soft robots and perform trajectory optimization with analytic gradients. \citet{takahashi2021differentiable} differentiate the simulation of fluids with solid coupling.

Many exciting applications of differentiable physics have been explored~\cite{Spielberg2019,Heiden2019,heiden2020NeuralSim,Wang2020first,murthy2021gradsim}. \citet{huang2021plasticinelab} propose a soft-body manipulation benchmark for differentiable physics. \citet{Toussaint2018differentiable} manipulate tools with the help of differentiable physics.
\citet{Song2020Identifying} perform system identification by learning from the trajectory, and \citet{Learning2020Song} then use the estimated frictional properties to help robotic pushing. 

A related line of work concerns the development of powerful automatic differentiation tools that can be used to differentiate simulation. DiffTaichi \cite{Hu2019:ICLR} provides a new programming language and a compiler, enabling the high-performance Taichi simulator to compute the gradients of the simulation. JAX MD~\cite{jaxmd2019} makes use of the JAX autodiff library~\cite{jax2018github} to differentiate molecular dynamics simulation. These works have specifically made use of vectorization on both CPU and GPU to achieve high performance on grids and particle sets, where the simulation is intrinsically parallel. In contrast, the simulation of articulated bodies is far more serial, as dynamics propagates along kinematic paths rather than acting in parallel on grid cells or particles.

TinyDiffSim~\cite{Coumans2020tiny} provides a templatized simulation framework that can leverage existing autodiff tools such as CppAD~\cite{Bradley2018cppad}, Ceres~\cite{ceres-solver}, and PyTorch~\cite{Paszke2019pytorch} to differentiate simulation. However, these methods introduce significant overhead in tracing the computation graph and accumulate substantial computational and memory costs when applied to articulated bodies.

Our method does not need to store the entire computation graph to compute the gradients.
We store the initial states of each simulation step and reproduce the intermediate results when needed by the backward pass. This checkpointing strategy is also used in training neural ODEs~\cite{zhuang2020adaptive} and large neural networks \cite{chen2016training,Gruslys2016memory,kirisame2021dynamic,shah2021memory}.
We are the first to adapt this technique to articulated dynamics, achieving dramatic reductions in memory consumption and enabling stable simulation of long experiences.

\mypara{Adjoint method.}
The adjoint method has been applied to fluid control~\cite{mcnamara2004fluid}, PDEs~\cite{Holl2020}, light transport simulation~\cite{david2020radiative}, and neural ODEs~\cite{chen2018neural,zhuang2020adaptive}. Recently, \citet{Geilinger2020add} proposed to use the adjoint method in multi-body dynamics. 
However, they operate in maximal coordinates and model body attachments using springs. This does not enforce physical validity of articulated bodies. In contrast, we operate in reduced coordinates and derive the adjoints for the articulated body algorithm and spatial algebra operators that properly model the body's joints. This supports physically correct simulation with joint torques, limits, Coriolis forces, and proper transmission of internal forces between links.



\mypara{Neural approximation.}
A number of works approximate physics simulation using neural networks~\cite{battaglia2016interaction,chang2017compositional,Mrowca2018,Schenck2018,Sanchez-Gonzalez2018,Sanchez-Gonzalez2020,Li2019:ICRA,belbute_peres_cfdgcn_2020,Ummenhofer2020,wandel2021learning,pfaff2021learning}. Physical principles have also been incorporated in the design of neural networks~\cite{Schutt2017,Anderson2019,Bogatskiy2020lorentz,Chen2020symplectic,cranmer2020lagrangian}. Approximate simulation by neural networks is naturally differentiable, but the networks are not constrained to abide by the underlying physical dynamics and simulation fidelity may degenerate outside the training distribution.


\begin{figure*}
    \centering
    \includegraphics[width=1\linewidth]{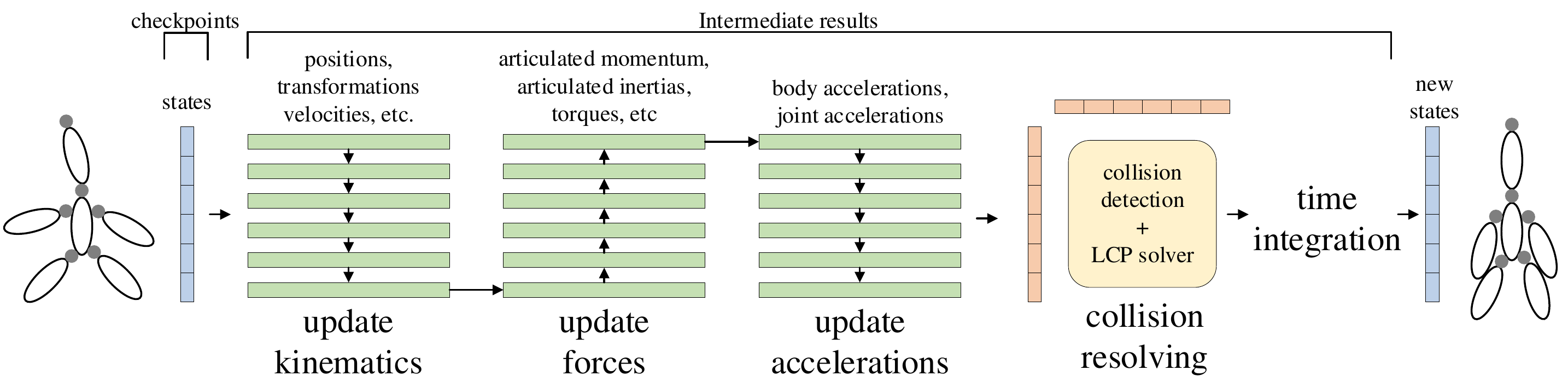}
    \vspace{-1.5em}
    \caption{{\bf The workflow of a simulation step.} Assume there are $n_q=6$ DoF. The initial state is an $n_q\times 3$ matrix containing the position, velocity, and control input of each joint. The forward dynamics will traverse the articulated body sequentially three times. This process is difficult to parallelize and will generate a large number of intermediate results. After the forward dynamics there is a collision resolution step with collision detection and an iterative Gauss-Seidel solver. The size of the initial state (position, velocity, and control input) is much smaller than that of intermediate results but the initial state has all the information to resume all the intermediate variables. }
    \label{fig:arti}
\end{figure*}

\section{Preliminaries}

\mypara{Articulated body dynamics.}
For the forward simulation, we choose the recursive Articulated Body Algorithm (ABA)~\cite{Featherstone2007RigidBD}, which has $O(n)$ complexity and is widely used in articulated body simulators~\cite{todorov2012mujoco,Coumans2015bullet,lee2018dart}. In each simulation step $k$, the states $\xx^k$ of the system consist of configurations in generalized coordinates and their velocities $\xx^k = [\qq^{k},\Dot{\qq}^{k}]\in \mathbb{R}^{2 n_q\times 1}$, where $n_q$ is the number of degrees of freedom in the system. Assuming each joint can be actuated by a torque, there is an $n_q$-dimensional control signal $\uu^k\in \mathbb{R}^{n_q\times 1}$. The discretized dynamics at this step can be written as $\ff^{k+1}(\uu^k,\xx^{k},\xx^{k+1})=0$.

\mypara{Adjoint method.}
We can concatenate the states in the entire time sequence into ${\xx=[\xx^{1},\xx^{2},...,\xx^{n_t}]\in \mathbb{R}^{n_t \cdot 2n_q\times 1}}$, with corresponding control input ${\uu=[\uu^{1},\uu^{2},...,\uu^{n_t}]\in \mathbb{R}^{n_t \cdot n_u\times 1}}$, where $n_t$ is the number of simulation steps. The concatenated dynamics equations can be similarly written as $\ff=[\ff^1,\ff^2,...,\ff^{n_t}]=0$. For a learning or optimization problem, we would usually define a scalar objective function $\Phi(\xx,\uu)$. To get the derivative of this function w.r.t. the control input $\uu$, one needs to compute
\begin{equation}
    \frac{d\Phi}{d\uu}=\PD{\Phi}{\uu}+\PD{\Phi}{\xx}\frac{d\xx}{d\uu}.
    \label{eq:diff_u}
\end{equation}
$\PD{\Phi}{\uu}$ is easy to compute for each single simulation step. But it is prohibitively expensive to directly solve $\PD{\Phi}{\xx}\frac{d\xx}{d\uu}$ because $\frac{d\xx}{d\uu}$ will be a $2n_qn_t\times n_qn_t$ matrix. 

Instead, we differentiate the dynamics $\ff(\xx,\uu)=0$ and obtain the constraint $\PD{\ff}{\xx}\frac{d\xx}{d\uu}=-\PD{\ff}{\uu}$. Applying the adjoint method~\cite{giles2000introduction}, $\PD{\Phi}{\xx}\frac{d\xx}{d\uu}$ equals to
\begin{equation}
  -\RR\trans\PD{\ff}{\uu} \textrm{ \hspace{4mm} such that \hspace{4mm} } \left(\PD{\ff}{\xx}\right)\trans\RR=\left(\PD{\Phi}{\xx}\right)\trans
  \label{eq:adj_back}
\end{equation}
Since the dynamics equation for one step only involves a small number of previous steps, the sparsity of $\PD{\ff}{\uu}$ and $\PD{\ff}{\xx}$ makes it easier to solve for the variable $\RR$ in Eq.~\ref{eq:adj_back}.

$\PD{\Phi}{\uu}-\RR\trans\PD{\ff}{\uu}$ is called the adjoint of $\uu$ and is equivalent to the gradient by our substitution. In the following derivation, we denote the adjoint of a variable $\sss$ by $\overline{\sss}$, and the adjoint of a function $f(\cdot)$ by $\overline{f}(\cdot)$.

\section{Efficient Differentiation}

In this section, we introduce our algorithm for the gradient computation in articulated body simulation. For faster differentiation, the gradients of articulated body dynamics are computed by the adjoint method. We derive the adjoint of time integration, contact resolution, and forward dynamics in reverse order. We then adapt the checkpoint method to articulated body simulation to reduce memory consumption.

\subsection{Adjoint Method for Articulated Dynamics}
\label{sec:adj}

One forward simulation step can be split into five modules, as shown in Figure~\ref{fig:arti}: perform the forward kinematics from the root link to the leaf links, update the forces from the leaves to the root, update the accelerations from the root to the leaves, detect and resolve collisions, and perform time integration. Backpropagation proceeds through these modules in reverse order.

\mypara{Time integration.}
Backpropagation starts from the time integration. As an example, for a simulation sequence with $n_t=3$ steps and an integrator with a temporal horizon of 2, the constraints $(\PD{\ff}{\xx})\trans\RR=(\PD{\Phi}{\xx})\trans$ in Equation~\ref{eq:adj_back} can be expanded as
\begin{equation}
\left[
\begin{array}{ccc}
(\PD{\ff^1}{\xx^1})\trans & (\PD{\ff^2}{\xx^1})\trans & 0  \\
0 & (\PD{\ff^2}{\xx^2})\trans & (\PD{\ff^3}{\xx^2})\trans  \\
0 & 0 & (\PD{\ff^3}{\xx^3})\trans
\end{array}\right]
\left[
\begin{array}{c}
\RR^1\\
\RR^2\\
\RR^3
\end{array}\right]=
\left[
\begin{array}{c}
(\PD{\Phi}{\xx^1})\trans\\
(\PD{\Phi}{\xx^2})\trans\\
(\PD{\Phi}{\xx^3})\trans
\end{array}\right]
\label{eq:adj_constraints}
\end{equation}
If using the explicit Euler integrator, we have $\PD{\ff^k}{\xx^k}=\II$. Initially $\RR^{n_t}=\PD{\Phi}{\xx^{n_t}}$. The following $\RR^k$ can be computed iteratively by 
\begin{equation}
    \RR^k=(\PD{\Phi}{\xx^k})\trans-(\PD{\ff^{k+1}}{\xx^k})\trans\RR^{k+1}, k=1,2,..,n_t-1.
\label{eq:adj_r}
\end{equation}
When $\RR^k$ backpropagates through time, the gradients $\overline{\uu^k}$ can also be computed by Equation~\ref{eq:adj_back}. In fact, by the way we calculate $\RR^k$, it equals the gradients of $\xx^k$. Other parameters can also be computed in a similar way as $\overline{\uu^k}$.  

\mypara{Collision resolution.} 
The collision resolution step consists of collision detection and a collision solver. Upon receiving the gradients $\overline{\xx}$ from the time integrator, the collision solver needs to pass the gradients to detection, and then to the forward dynamics. 
In our collision solver, we construct a Mixed Linear Complementarity Problem (MLCP)~\cite{Stepien2013}:
\begin{equation}
\begin{split}
    &\aa = \AA\xx+\bb\  \\
    \textrm{ s.t. \hspace{2mm} } \bm{0}\leq &\aa \perp \xx\geq \bm{0}\textrm{\hspace{1mm} and \hspace{1mm}} \cc_- \leq \xx  \leq\cc_+ ,
\end{split}
\label{eq:mlcp}
\end{equation}
where $\xx$ is the new collision-free state, $\AA$ is the inertial matrix, $\bb$ contains the relative velocities, and $\cc_-, \cc_+$ are the lower bound and upper bound constraints, respectively. We use the projected Gauss-Seidel (PGS) method to solve this MLCP. This iterative solve trades off accuracy for speed such that the constraints $\aa\trans\xx=0$ might not hold on termination. In this setting, where the solution is not guaranteed to satisfy constraints, implicit differentiation~\cite{Belbute2018,Liang2019,Qiao2020Scalable} no longer works. Instead, we design a reverse version of the PGS solver using the adjoint method to compute the gradients. Further details are in the supplement. In essence, the solver mirrors PGS, substituting the adjoints for the original operators.
This step passes the gradients from the collision-free states to the forward dynamics.

\mypara{Forward dynamics.}
Our articulated body simulator propagates the forward dynamics as shown in the green blocks of Figure~\ref{fig:arti}. Each operation in the forward simulation has its corresponding adjoint operation in the backward pass. To compute the gradients, we derive adjoint rules for the operators in spatial algebra~\cite{Featherstone2020spatial}.

As a simple example, a spatial vector $\pp_i=[\ww,\vv]\in\mathbb{R}^6$ representing the bias force of the $i$-th link is the sum of an external force $[\ff_1, \ff_2]\in\mathbb{R}^6$ and a cross product of two \ylq{spatial motion vectors} $[\ww_1,\vv_1], [\ww_2,\vv_2]\in\mathbb{R}^6$:
\begin{equation}
\left[
\begin{array}{c}
\ww\\
\vv
\end{array}\right]=
\left[
\begin{array}{c}
\ff_1+\ww_1 \times \ww_2+\vv_1\times\vv_2\\
\ff_2+\ww_1\times\vv_2
\end{array}\right].
\label{eq:sm-crossf}
\end{equation}
Once we get the adjoint $[\overline{\ww}, \overline{\vv}]$ of $\pp_i=[\ww,\vv]$, we can propagate it to its inputs:
\begin{equation}
\begin{split}
\left[
\begin{array}{c}
\adj{\ww}{1}\\
\adj{\vv}{1}
\end{array}\right]&=
\left[
\begin{array}{c}
-\adj{\ww}\times\ww_2-\adj{\vv}\times\vv_2 \\
-\adj{\ww}\times\vv_2
\end{array}\right] \\
\left[
\begin{array}{c}
\adj{\ww}{2}\\
\adj{\vv}{2}
\end{array}\right]&=
\left[
\begin{array}{c}
\adj{\ww}\times\ww_1\\
\adj{\ww}\times\vv_1+\adj{\vv}\times\ww_1
\end{array}\right]
,
\left[
\begin{array}{c}
\adj{\ff}{1}\\
\adj{\ff}{2}
\end{array}\right]=
\left[
\begin{array}{c}
\overline{\ww}\\
\overline{\vv}
\end{array}\right] \\
\end{split}
\label{eq:adjdemo}
\end{equation}
This example shows the adjoint of one forward operation. The time and space complexity of the original operation and its adjoint are the same as the forward simulation. The supplement provides more details on the adjoint operations.

\subsection{Checkpointing for Articulated Dynamics}

The input and output variables of one simulation step have relatively small dimensionalities (positions, velocities, and accelerations), but many more intermediate values are computed during the process. Although this is not a bottleneck for forward simulation because the memory allocation for intermediate results can be reused across time steps, it becomes a problem for differentiable simulation, which needs to store the entire computation graph for the backward pass. This causes memory consumption to explode when simulation proceeds over many time steps, as is the case when small time steps are needed for accuracy and stability, and when the effects of actions take time to become apparent.

\begin{figure}
    \centering
    \includegraphics[width=1\linewidth]{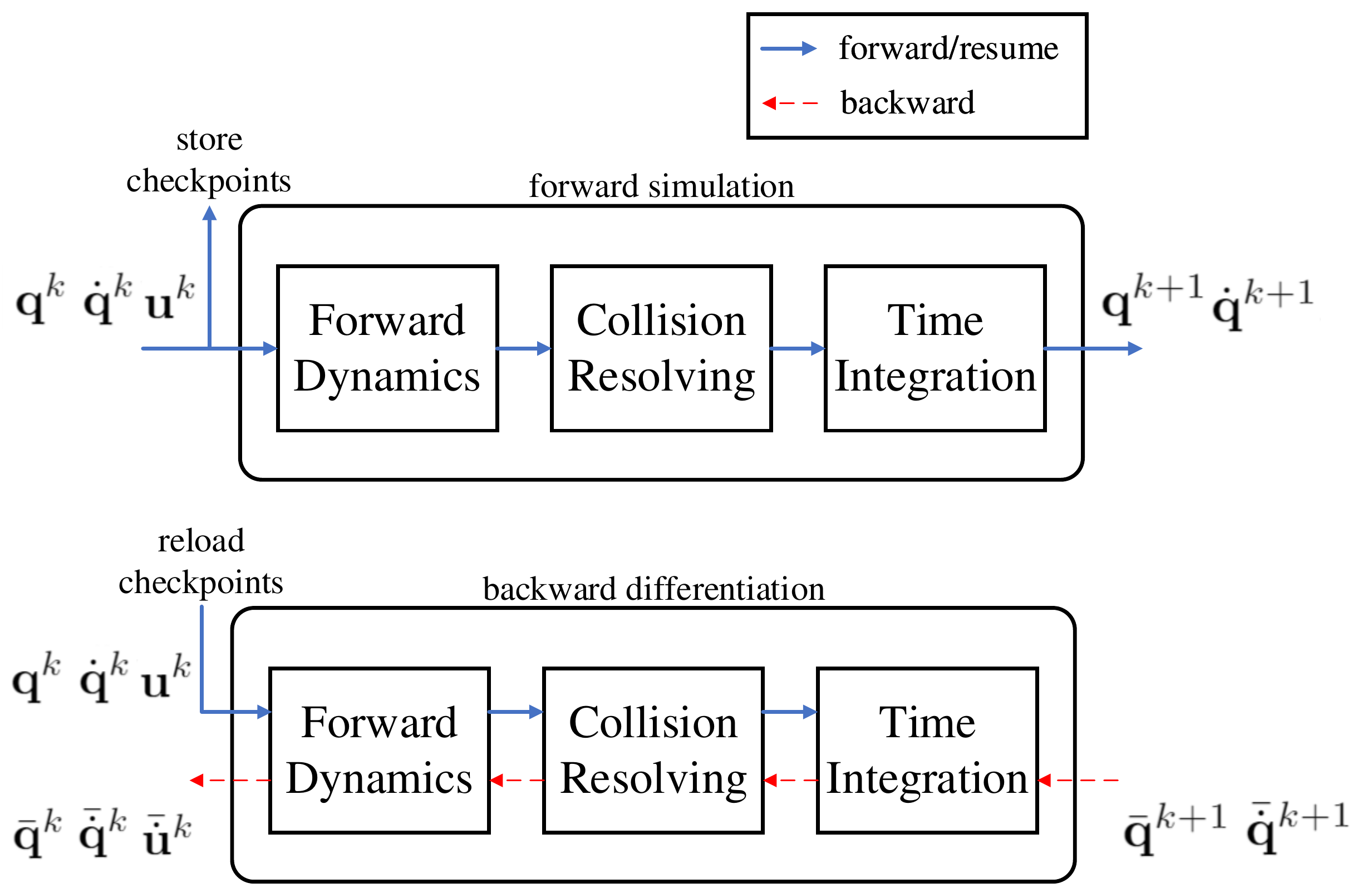}
    \vspace{-1em}
    \caption{{\bf Differentiation of articulated body dynamics.} {\em Top:} forward simulation at step $k$. The simulator stores the states $\qq^k, \Dot{\qq}^k$, and control input $\uu^k$. {\em Bottom:} During backpropagation, the checkpoints are reloaded and the simulator runs one forward step to reconstruct the intermediate variables.
    Beginning with the gradients from step $k+1$, we use the adjoint method to sequentially compute the gradients of all variables through time integration, the constraint solver, and forward dynamics.
    }
    \label{fig:fb}
\end{figure}

The need for scaling to larger simulation steps motivates our adaptation of the checkpointing method~\cite{GriewankWalther2000,chen2016training}.
Instead of saving the entire computation graph, we choose to store only the initial states in each time step.
We use this granularity because (a) the states of each step have the smallest size among all essential information sufficient to replicate the simulation step, (b) finer checkpoints are not useful because at least one step of intermediate results needs to be stored in order to do the simulation, and (c) sparser checkpoints will use less memory but require multiple steps for reconstructing intermediate variables, costing more memory and computation. We validate our checkpointing policy with an ablation study in the supplement.
Figure~\ref{fig:fb} illustrates the scheme. During forward simulation, we store the simulation state in the beginning of each time step. During backpropagation, we reload the checkpoints (blue arrows) and redo the forward (one-step) simulation to generate the computation graph, and then compute the gradients using the adjoint method in reverse order (red arrows).


In summary, assume the simulation step consists of two parts:
\begin{equation}
    \ZZ^k=G(\xx^{k-1}),\qquad \xx^k=F(\ZZ^k),
\label{equ:intermediate}
\end{equation}
where $\ZZ^k$ represents all the intermediate results.
After each step, we free the space consumed by $\ZZ^k$, only storing $\xx^k$.
During backpropagation, we recompute  $\ZZ^k$ from $\xx^{k-1}$ and use the adjoint method to compute gradients of the previous step:
\begin{eqnarray}
    \ZZ^k&=&G(\xx^{k-1}), \\
    \overline{\ZZ^k}&=&\overline{F}(\ZZ^k, \overline{\xx^k}), \\
    \overline{\xx_{k-1}}&=&\overline{G}(\xx_{k-1},\overline{\ZZ^k}).
\end{eqnarray}

\section{Reinforcement Learning}
\label{sec:rl}
Our method can be applied to simulation-based reinforcement learning (RL) tasks to improve policy performance as well as convergence speed.
By computing gradients with respect to the actions, differentiable simulation can provide more information about the environment.
We suggest two approaches to integrating differentiable physics into RL.

\mypara{Sample enhancement.}
We can make use of the computed gradients to generate samples in the neighborhood around an existing sample.
Specifically, given a sample $(s,a_0,s_0',r_0)$ from the history, with observation $s$, action $a_0$, next-step observation $s_0'$, and reward $r_0$, we can generate new samples $(s,a_k,s_k',r_k)$ using first-order approximation:
\begin{eqnarray*}
    \aa_k&=&\aa_0+\Delta \aa_k, \\
    \sss_k'&=&\sss_0'+\frac{\partial \sss_0'}{\partial \aa_0}\Delta \aa_k, \\
    r_k&=&r_0+\frac{\partial r_0}{\partial \aa_0}\Delta \aa_k,
\end{eqnarray*}
where $\Delta \aa_k$ is a random perturbation vector.
This method, which we call {\em sample enhancement}, effectively generates as many approximately accurate samples as desired for learning purposes.
By providing a sufficient number of generated samples around the neighborhood, the critic can have a better grasp of the function shape (patchwise rather than pointwise), and can thus learn and converge faster.

\mypara{Policy enhancement.}
Alternatively, the policy update can be adjusted to a format compatible with differentiable physics, usually dependent on the specific RL algorithm.
For example, in MBPO~\cite{janner2019trust}, the policy network is updated using the gradients of the critic:
\begin{equation}
    \lL_{\mu}=-Q(\sss,\mu(\sss))+Z,
    \label{equ:trueloss}
\end{equation}
where $\lL_{\mu}$ is the loss for the policy network $\mu$, $Q$ is the value function for state-action pairs, and $Z$ is the regularization term.
To make use of the gradients, we can expand the Q function one step forward, 
\begin{gather}
    \frac{\partial Q(\sss,\aa)}{\partial \aa}=\frac{\partial r}{\partial \aa}+\gamma\frac{\partial Q(\sss',\mu(\sss'))}{\partial \sss'}\frac{\partial \sss'}{\partial \aa},
    \label{equ:fakelossgrad}
\end{gather}
and substitute it in Eq.~\ref{equ:trueloss}:
\begin{gather}
    \lL_{\mu}'=-\frac{\partial Q(\sss,\aa)}{\partial \aa}\mu(\sss)+Z.
    \label{equ:fakeloss}
\end{gather}
Eqs.~\ref{equ:trueloss} and \ref{equ:fakeloss} yield the same gradients with respect to the action, which provide the gradients of the network parameters.
This method, which we call {\em policy enhancement}, constructs the loss functions with embedded ground-truth gradients so that the policy updates are physics-aware and generally more accurate than merely looking at the Q function itself, thus achieving faster convergence and even potentially higher reward.

\section{Results}
\label{sec:experiments}
\begin{table}[t]
\newcolumntype{Z}{S[table-format=4.1,table-auto-round]}
\centering
\ra{1.1}
\resizebox{1\linewidth}{!}{
	\small
	\begin{tabular}{l|ZZZZZ}
		\toprule
		steps & 50 & 100  & 500 & 1000 & 5000   \\
		\hline
		Ceres    &  100.0  & 200.0  & 1100.0 & 2700.0 & 23600.0\\
		CppAD    &  16.0  & 26.0  & 160.0 & 320.0 & 3100.0  \\
	    JAX   & 200.0 & 200.0  & 400.0 & 700.0 & 3000.0  \\
	    PyTorch   &  1200.0 & 2400.0  & 11700.0 & 12400.0 & N/A  \\
		\hline
		Ours    &  \bfseries 0.3 &\bfseries 0.3  &\bfseries 0.7 &\bfseries 1.2 &\bfseries 5.0 \\
		\bottomrule
	\end{tabular}
}
\vspace{-2mm}
\caption{{\bf Peak memory use of different simulation frameworks.} 
The memory footprint (MB) of our framework is more than two orders of magnitude lower than autodiff methods. PyTorch crashes at 5,000 simulation steps. ADF fails to compute the gradients in a reasonable time (10 min) and is not included here for this reason.
} 
\label{tab:all_memory}
\end{table}

\begin{table}[t]
\newcolumntype{Z}{S[table-format=4.1,table-auto-round]}
\centering
\ra{1.1}
\resizebox{1\linewidth}{!}{
	\small
	\begin{tabular}{l|ZZZZZ}
		\toprule
		steps & 50 & 100  & 500 & 1000 & 5000   \\
		\hline
		ADF    & 25.7   & 25.5  & 25.1 & 32.1 & 58.4  \\
		Ceres    &  27.2  & 27.5  & 27.2 & 34.0 & 58.2 \\
		CppAD    &  2.4  & 2.4  & 2.3 & 2.3 & 4.5  \\
	    JAX   &  53.3 & 46.1  & 43.1 & 42.7 & 42.3  \\
	    PyTorch   &  195.6 & 192.2  & 199.2 & 192.8 & N/A  \\
	    \hline
		Ours    & \bfseries  0.3 &\bfseries  0.3  &\bfseries  0.2 &\bfseries  0.2 &\bfseries  0.2 \\
		\bottomrule
	\end{tabular}
}
\vspace{-2mm}
\caption{{\bf Simulation time for forward pass.} Ours is at least an order of magnitude faster than autodiff tools (in msec). PyTorch crashes at 5,000 simulation steps. CppAD is the fastest baseline.
}
\label{tab:all_time}
\end{table}

\begin{table}[t]
\newcolumntype{Z}{S[table-format=4.1,table-auto-round]}
\centering
\ra{1.1}
\resizebox{1\linewidth}{!}{
	\small
	\begin{tabular}{l|ZZZZZ}
		\toprule
		steps & 50 & 100  & 500 & 1000 & 5000   \\
		\hline
		Ceres    &  29.5  & 29.6  & 29.5 & 37.6 & 62.8 \\
		CppAD    & 2.4 & 2.3  & 2.3 & 2.4 & 4.8  \\
	    JAX   &   148.1  & 125.7  & 127.6 & 129.6 & 126.6  \\
	    PyTorch   &  275.7 & 273.3  & 280.8 & 272.1 & N/A  \\
	    \hline
		Ours    & \bfseries  1.2 &\bfseries  1.1  &\bfseries  1.2 &\bfseries  1.1 &\bfseries  1.2 \\
		\bottomrule
	\end{tabular}
}
\vspace{-2mm}
\caption{{\bf Simulation time for the backward pass.} Ours is the fastest (in msec). PyTorch crashes at 5,000 simulation steps. 
}
\label{tab:backtime}
\end{table}

For experiments, we will first scale the complexity of simulated scenes and compare the performance of our method with autodiff tools. Then we will integrate differentiable physics into reinforcement learning and use our method to learn control policies. Lastly, we will apply differentiable articulated dynamics to solve motion control and parameter estimation problems.

\begin{figure}
\centering
\begin{tabular}{@{}c@{\hspace{1mm}}c@{\hspace{1mm}}c@{\hspace{1mm}}@{}}
    \includegraphics[width=.48\linewidth]{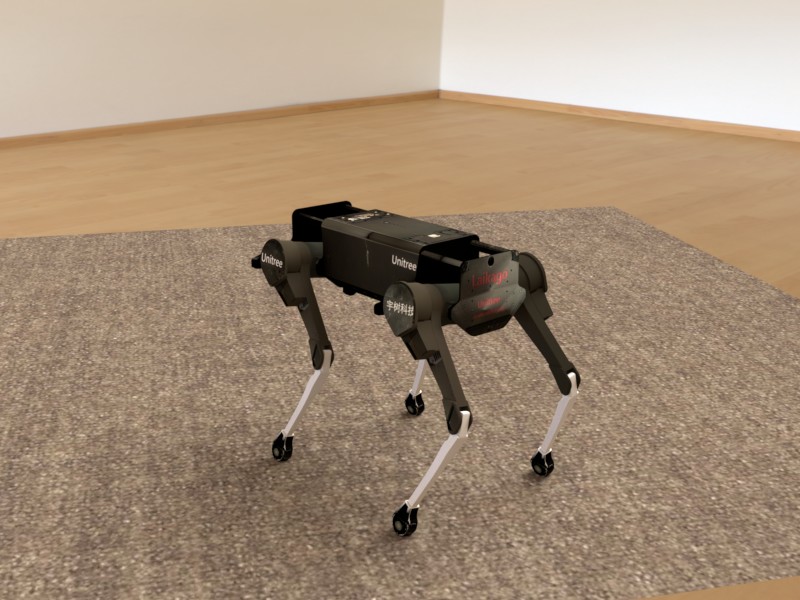} & \includegraphics[width=.48\linewidth]{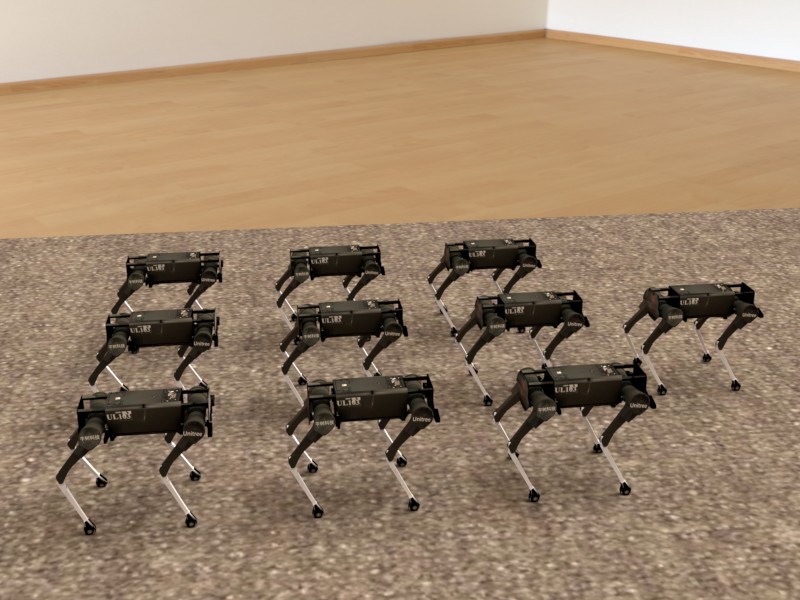} \\
    (a) One Laikago & (b) Ten Laikago robots \\
    \includegraphics[width=.45\linewidth]{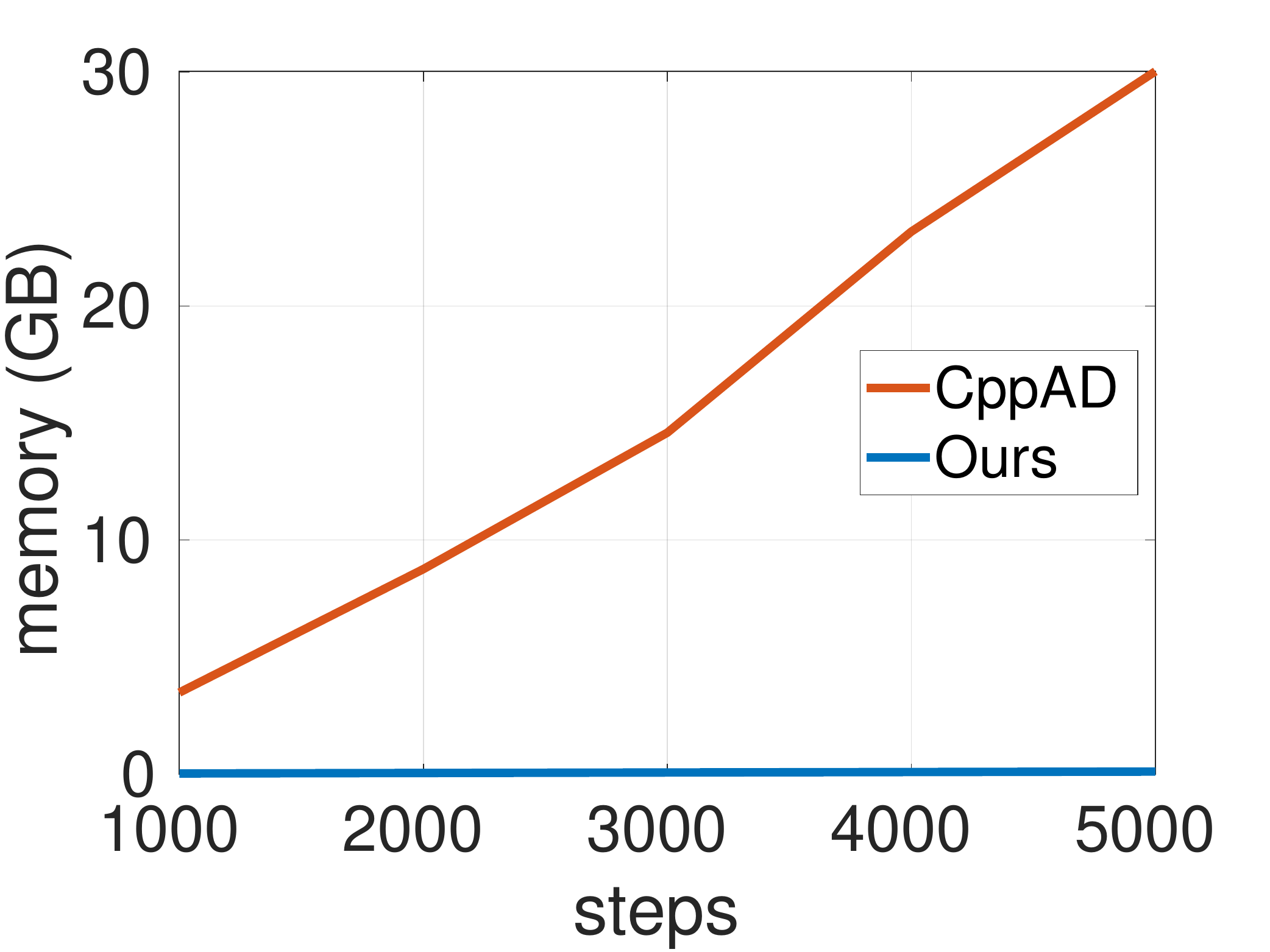} 
     & \includegraphics[width=.45\linewidth]{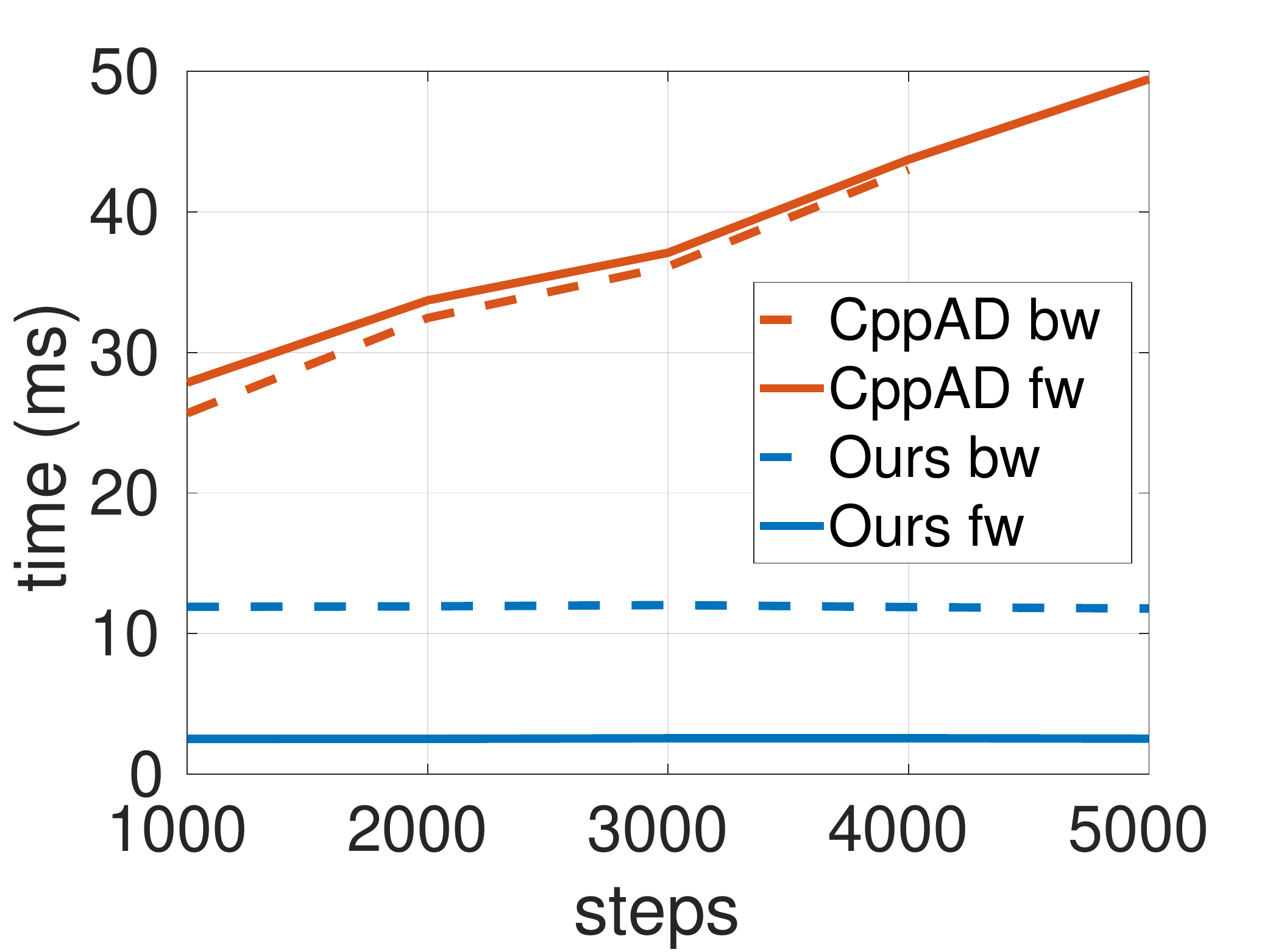}
    \\
     (c) Memory footprint  & (d) Time per step
\end{tabular}
\vspace{-2mm}
\caption{{\bf Memory and speed of our method vs.\ CppAD}, the fastest autodiff method. (a,b) Two scenes used in experiments. (c,d) Memory consumption and per-step runtime when simulating ten Laikago robots for different numbers of steps. CppAD crashes when simulating for 5,000 steps due to memory overflow. 
}
\label{fig:profile}
\end{figure}

\subsection{Comparison with Autodiff Tools}
\label{sec:autodiff}

Using existing autodiff tools is a convenient way to derive simulation gradients. However, these tools are not optimized for articulated dynamics, which adversely affects their computation and memory consumption in this domain.
We compare our method with state-of-the-art autodiff tools, including CppAD~\cite{Bradley2018cppad}, Ceres~\cite{ceres-solver}, PyTorch~\cite{Paszke2019pytorch}, autodiff~\cite{Leal2018Autodiff} (referred to as ADF to avoid ambiguity), and JAX~\cite{jax2018github}.
All our experiments are performed on a desktop with an Intel(R) Xeon(R) W-2123 CPU @ 3.60GHz with 32GB of memory.


\mypara{One robot}. In the first round, we profile all the methods by simulating one Laikago robot standing on the ground. Figure~\ref{fig:profile}(a) provides a visualization. The state vector of the Laikago has 37 dimensions (19-dimensional positions and 18-dimensional velocities).  We vary the number of simulation steps from 50 to 5,000. JAX is based on Python and its JIT compiler cannot easily be used in our C++ simulation framework. To test the performance of JAX, we write a simple approximation of our simulator in Python, with similar computational complexity as ours. The simplified test code can be found in the supplement.

The memory consumption of different methods is listed in Table~\ref{tab:all_memory}. The memory consumed by autodiff tools is orders of magnitude higher than ours. Among all the auto-differentiation tools, JAX and CppAD are the most memory-efficient. ADF fails to backpropagate the gradients in a reasonable time. We show in the supplement that the computation time of backpropagation in ADF is exponential in the depth of the computational graph. Note that articulated body simulation is intrinsically deep. In this experiment, the depth of one simulation step can reach $10^3$ due to sequential iterative steps in the forward dynamics and the contact solver. PyTorch crashes at 5,000 simulation steps because of memory overflow.

Table~\ref{tab:all_time} reports the average time for a forward simulation step. The numbers indicate that our method is faster than others by at least an order of magnitude. Our method has a constant time complexity per step. The time per step of ADF, Ceres, and CppAD increases with the number of steps. Note that JAX and Torch are optimized for heavily data-parallel workloads. Articulated body simulation is much more serial in nature and cannot take full advantage of their vectorization. Table~\ref{tab:backtime} reports the time for backward step. Our method is the fastest when computing the gradients. 

\mypara{Ten robots}. To further test the efficiency of our method, we simulate a scene with 10 Laikago robots. Figure~\ref{fig:profile}(b) provides a visualization.
In this experiment, our method is compared with CppAD, the best-performing autodiff framework according to the experiments reported in Tables~\ref{tab:all_memory} and~\ref{tab:all_time}. The number of simulation steps is varied from 1,000 to 5,000. Figure~\ref{fig:profile}(c) shows that the memory footprint of our method is negligible compared to CppAD. Our method only needs to store 5 KB of data per step, while CppAD needs to save the full data and topology of the computational graph. The forward and backward time per step are plotted in Figure~\ref{fig:profile}(d). Our backward pass takes longer than our forward pass because the method first runs forward simulation to reconstruct the intermediate variables. Nevertheless, our method is much faster than CppAD and the performance gap grows with simulation length.

\subsection{Integration with Reinforcement Learning}
\label{sec:rlenhance}
We demonstrate the improvements from the two RL enhancement methods described in Section~\ref{sec:rl}.
Note that we cannot use the two methods together because policy enhancement requires the gradients at the sample point, which the extra samples from sample enhancement do not have (unless higher-order gradients are computed).
Thus, we test the two techniques separately.
We use the model-based MBPO optimizer as the main baseline~\cite{janner2019trust}.
All networks use the PyTorch default initialization scheme.



\begin{figure}
\centering
\begin{tabular}{@{}c@{\hspace{0.5mm}}cc@{}}
    \includegraphics[width=.5\linewidth]{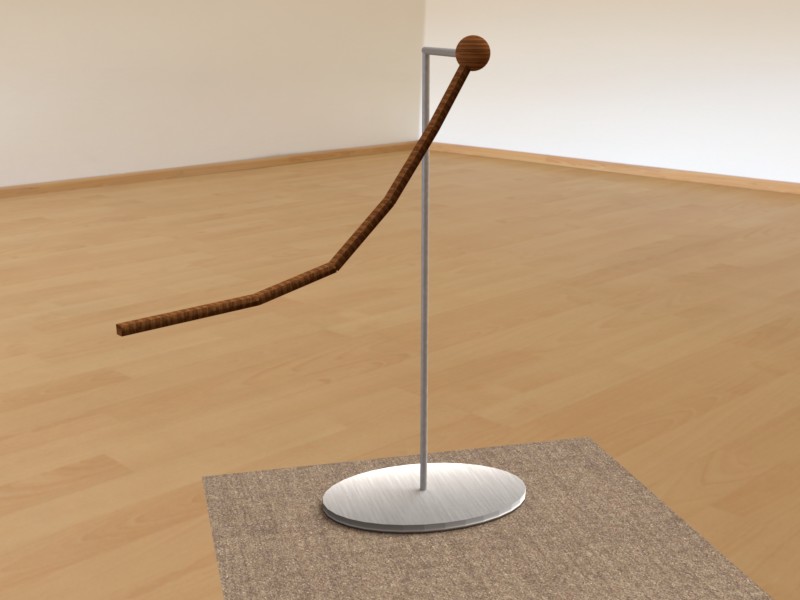} &
    \includegraphics[width=.5\linewidth]{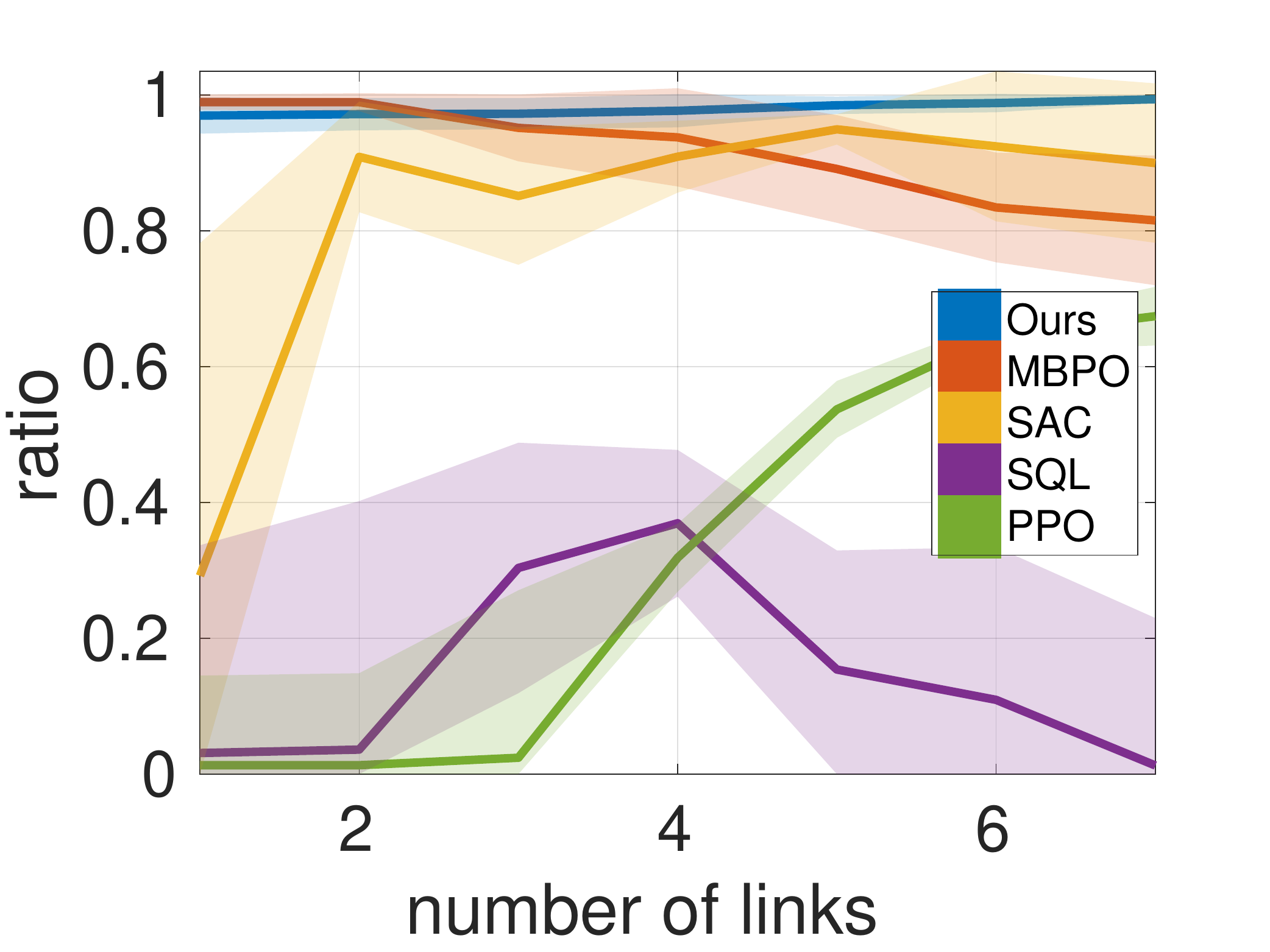} \\
    (a) Pendulum & (b) Relative reward \\
    \includegraphics[width=.5\linewidth]{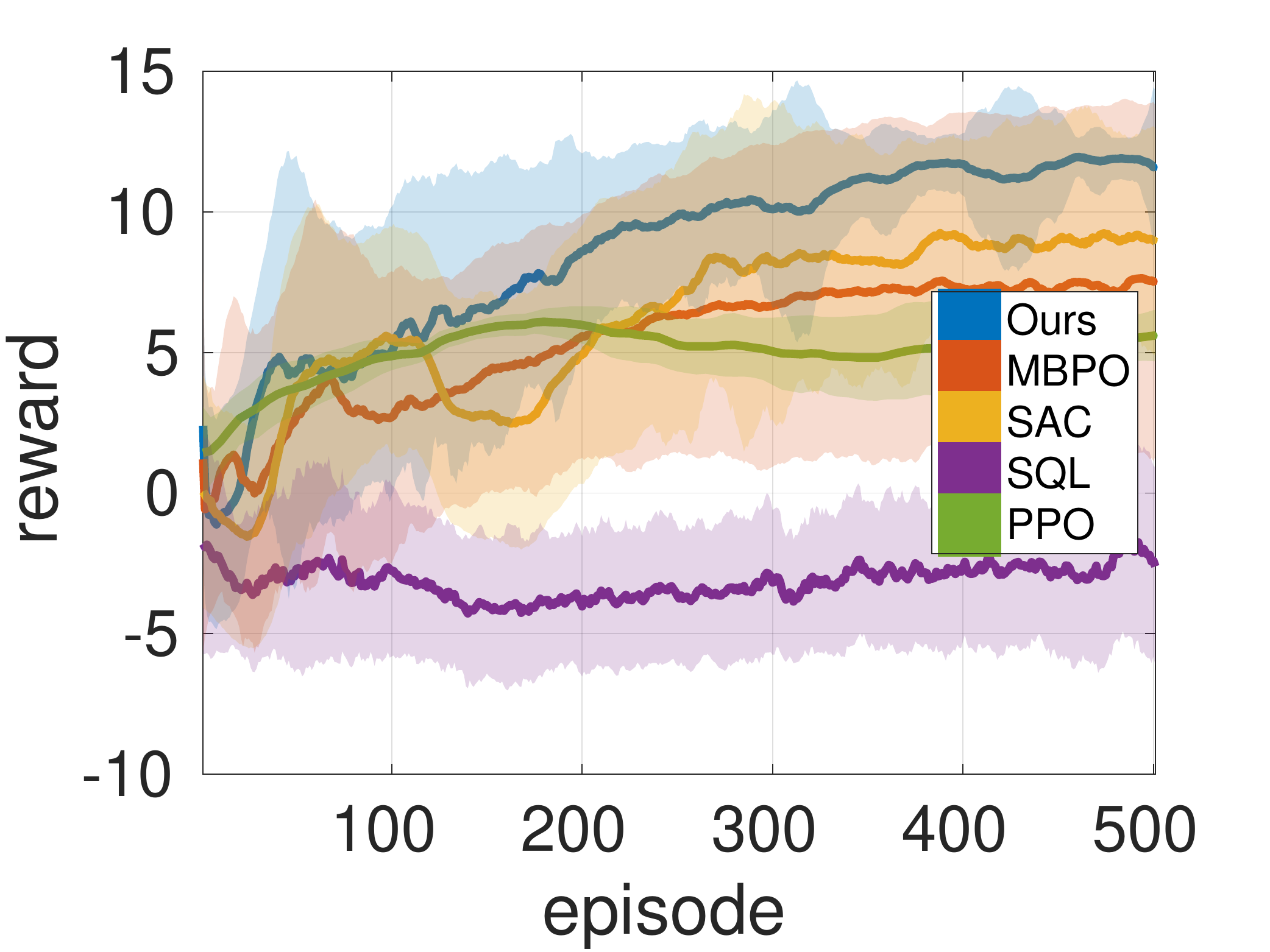}& 
    \includegraphics[width=.5\linewidth]{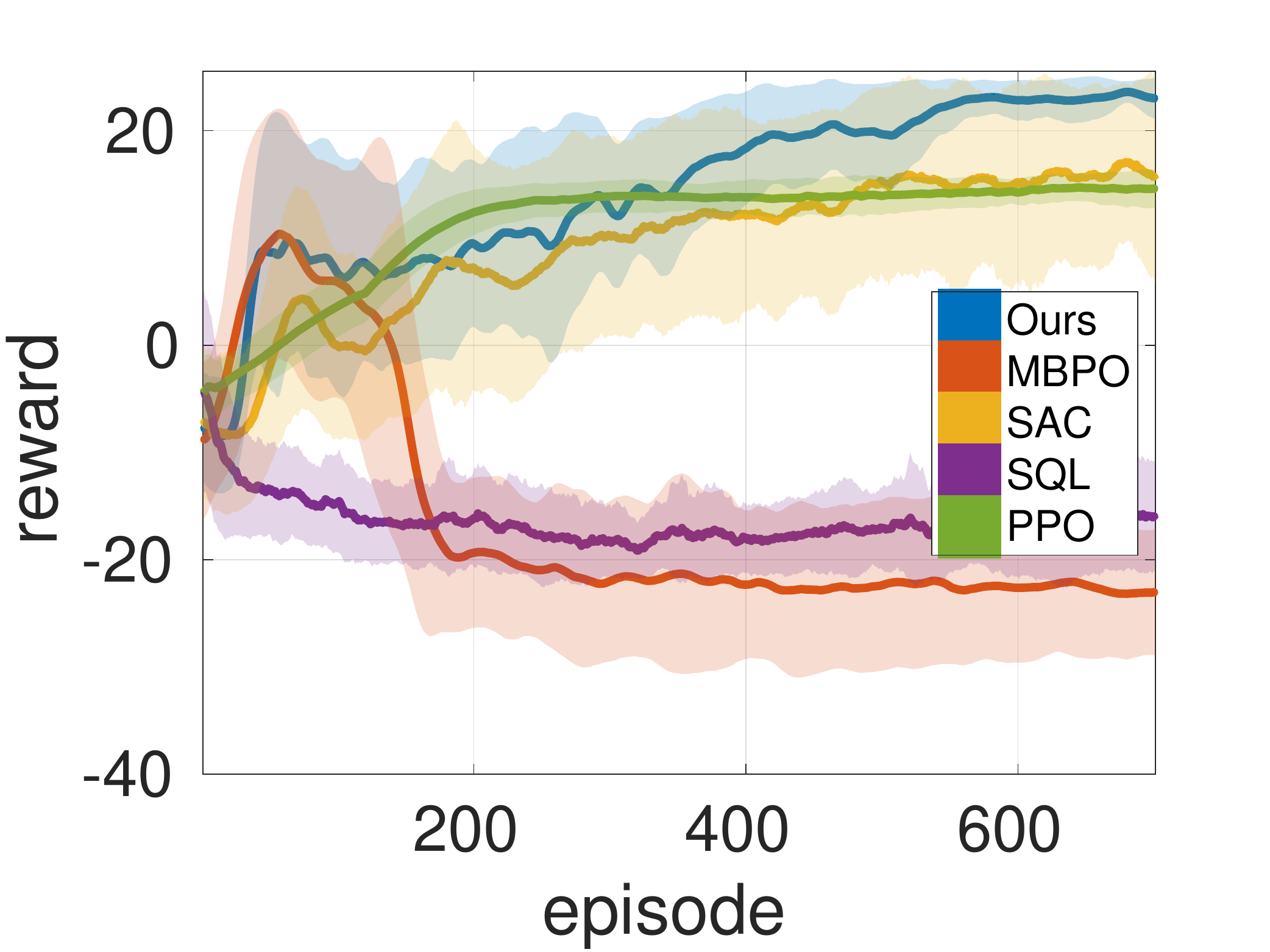}
    \\
    (c) 5-link learning curves & (d) 7-link learning curves
\end{tabular}
\vspace{-2mm}
\caption{{\bf The $n$-link pendulum task.}  Our method attains higher reward than MBPO.
}
\label{fig:pendulum}
\end{figure}

\mypara{N-link pendulum.} 
We first test our {\em policy enhancement} method in a simple scenario, where an $n$-link pendulum needs to reach a target point by applying torques on each of the joints under gravity (Figure~\ref{fig:pendulum}(a)).
The target point is fixed to be the highest point reachable, and initially the pendulum is positioned horizontally.
The reward function is the progress to the target between consecutive steps:
\begin{gather}
    r=\|\xx_t-\xx_g\|^2-\|\xx_{t-1}-\xx_g\|^2,
\end{gather}
where $\xx_t$ is the end-effector location at time $t$, and $\xx_g$ is the target location.

We trained each model for $100n$ epochs, where $n$ is the number of links.
The results are shown in Figure~\ref{fig:pendulum}.
In Figure~\ref{fig:pendulum}(b), we report the relative reward of each task, which is defined as the attained reward divided by the maximal possible reward.
MBPO works well in easy scenarios with up to 3 links, but its performance degrades starting at 4 links. 
In contrast, our model reaches close to the best possible reward for all systems.
The learning curves for the 5-link and 7-link systems are shown in Figure~\ref{fig:pendulum}(c,d).
MBPO does not attain satisfactory results for 6- and 7-link systems.
We observed that around the 100th epoch, the losses for the Q function of the MBPO method increased a lot.
We hypothesize that this is because the complexity of the physical system exceeds the expressive power of the model network in MBPO.


\begin{figure}
\centering
\begin{tabular}{@{}c@{\hspace{1mm}}c@{\hspace{1mm}}c@{}}
    \includegraphics[width=.48\linewidth]{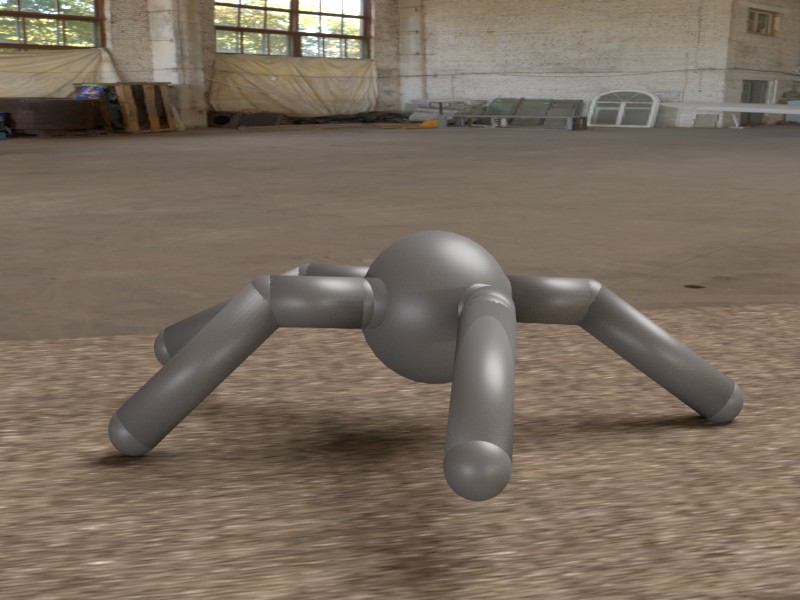} &
    \includegraphics[width=.48\linewidth]{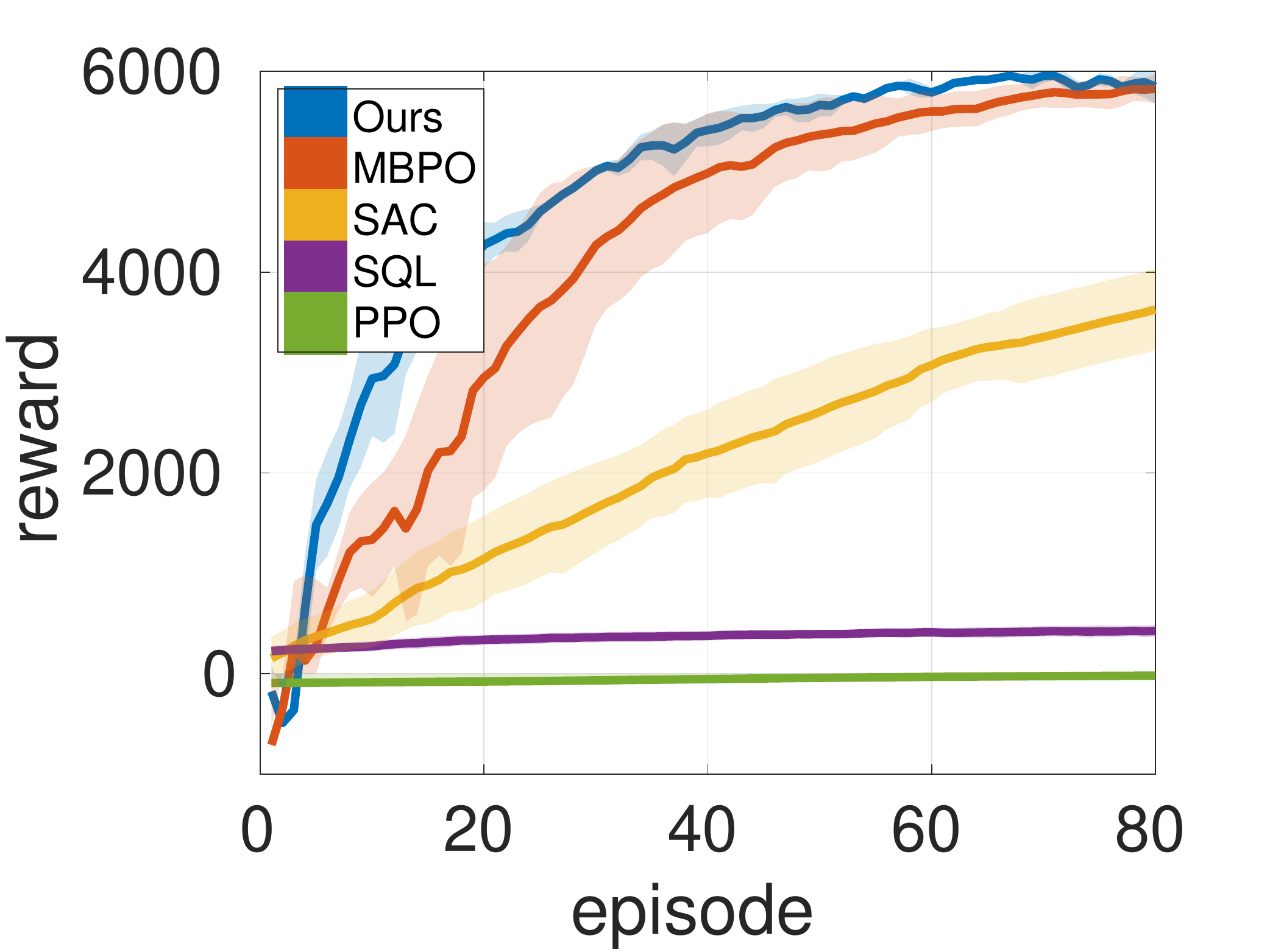} \\
    (a) Ant walking & (b) Reward curve
\end{tabular}
\vspace{-2mm}
\caption{{\bf The MuJoCo Ant task.}  Using our differentiable simulator to generate extra samples accelerates learning.}
\label{fig:ant}
\end{figure}

\mypara{MuJoCo Ant}. 
Next, we test our {\em sample enhancement} method on the MuJoCo Ant. In this scenario, a four-legged robot on the ground needs to learn to walk towards a fixed heading (Figure~\ref{fig:ant}(a)).
The scenario is the same as the standard task defined in MuJoCo, except that the simulator is replaced by ours.
We also compare with other methods (SAC~\cite{haarnoja2018soft}, SQL~\cite{haarnoja2017reinforcement}, and PPO~\cite{schulman2017proximal} implemented in Ray RLlib~\cite{liang2018rllib}) for reference.
For every true sample we get from the simulator, we generate 9 extra samples around it using our sample enhancement method.
Other than that, everything is the same as in MBPO.
We repeat the training 4 times for all methods. Figure~\ref{fig:ant}(b) shows the reward over time.
Our method exhibits faster convergence.


We can also transfer the policy trained in our differentiable simulator to MuJoCo. This tests the fidelity of our simulator and robustness of the learned policy to simulator details. The results are reported in the supplement.


\begin{figure}[t]
\centering
\begin{tabular}{@{}c@{\hspace{1mm}}c@{\hspace{1mm}}c@{}}
    \includegraphics[width=.48\linewidth]{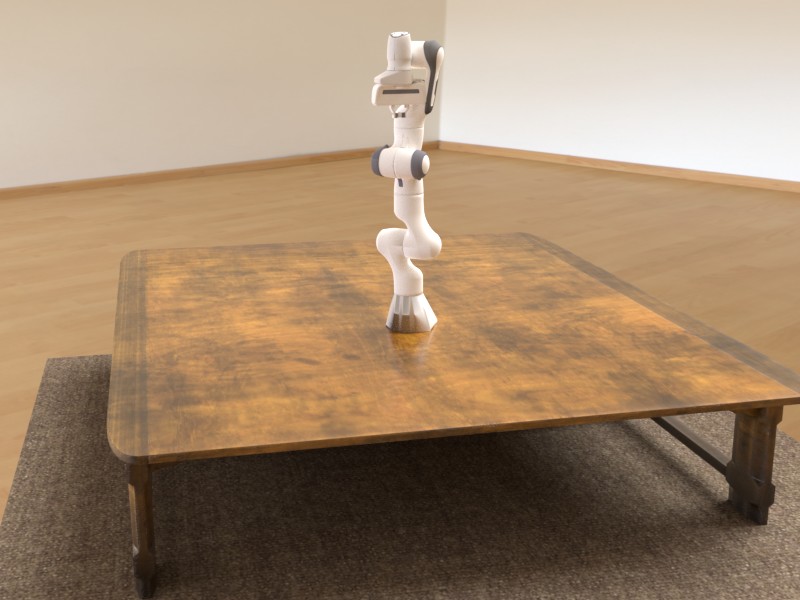} &
    \includegraphics[width=.48\linewidth]{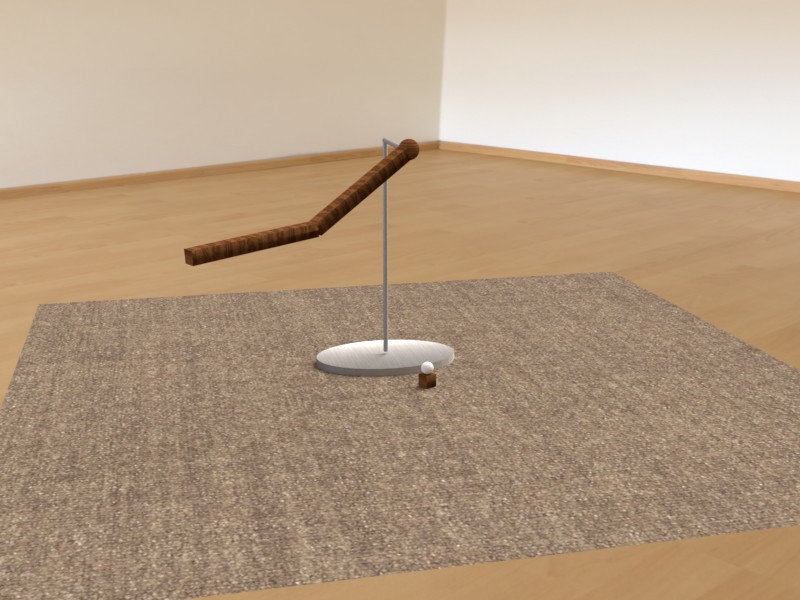} \\
    \includegraphics[width=.48\linewidth]{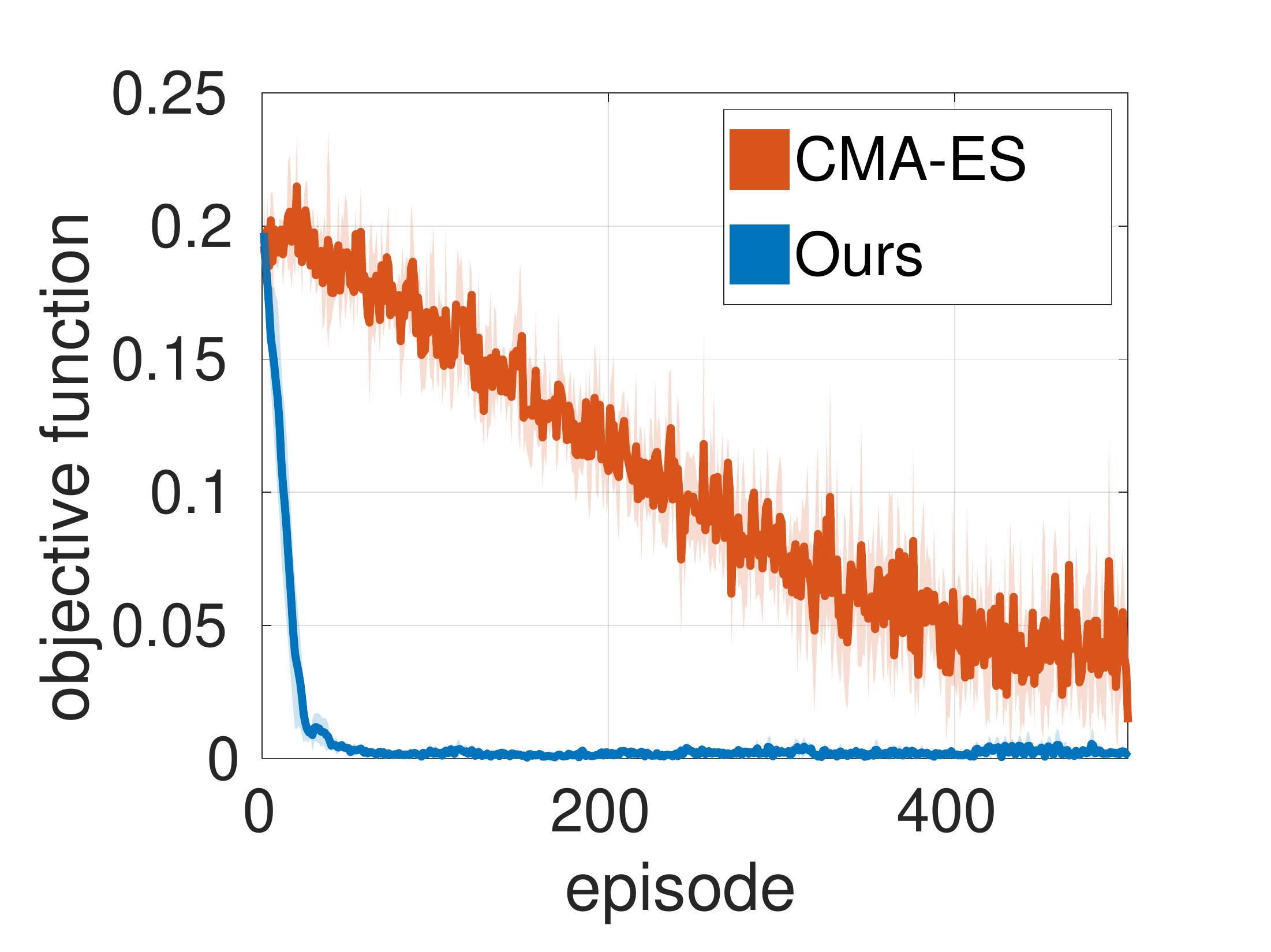} &
    \includegraphics[width=.48\linewidth]{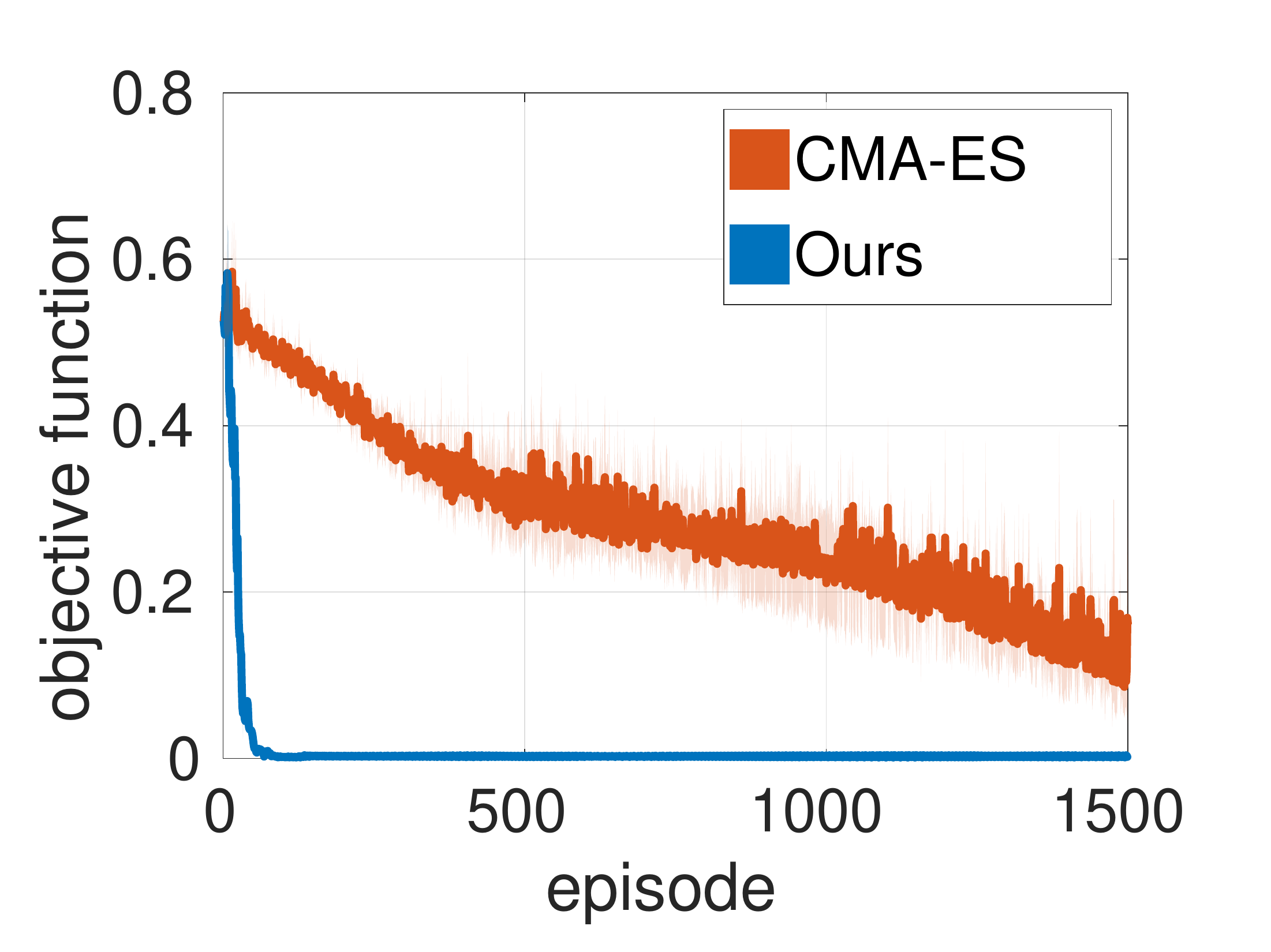} \\
    \includegraphics[width=.48\linewidth]{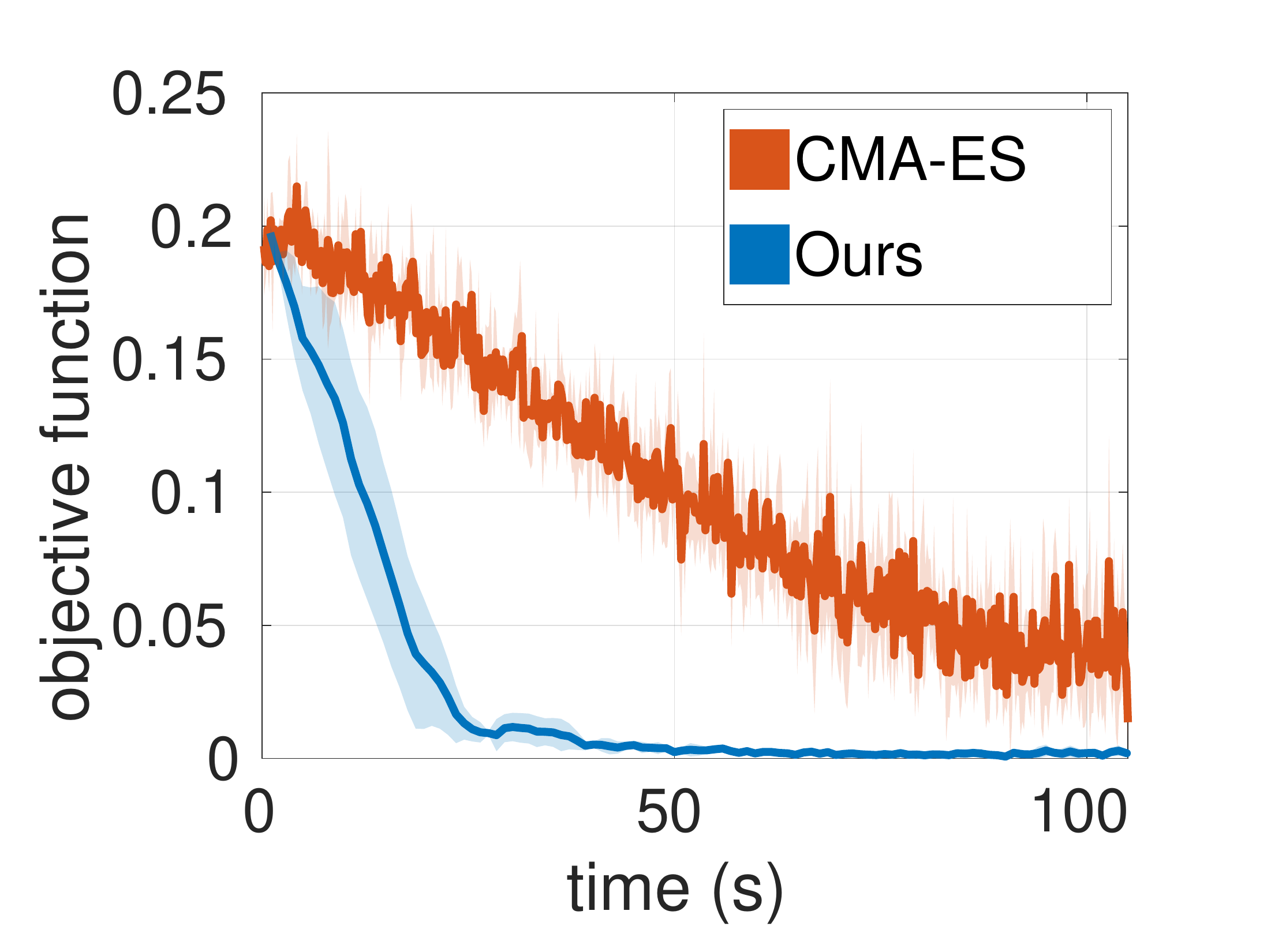} &
    \includegraphics[width=.48\linewidth]{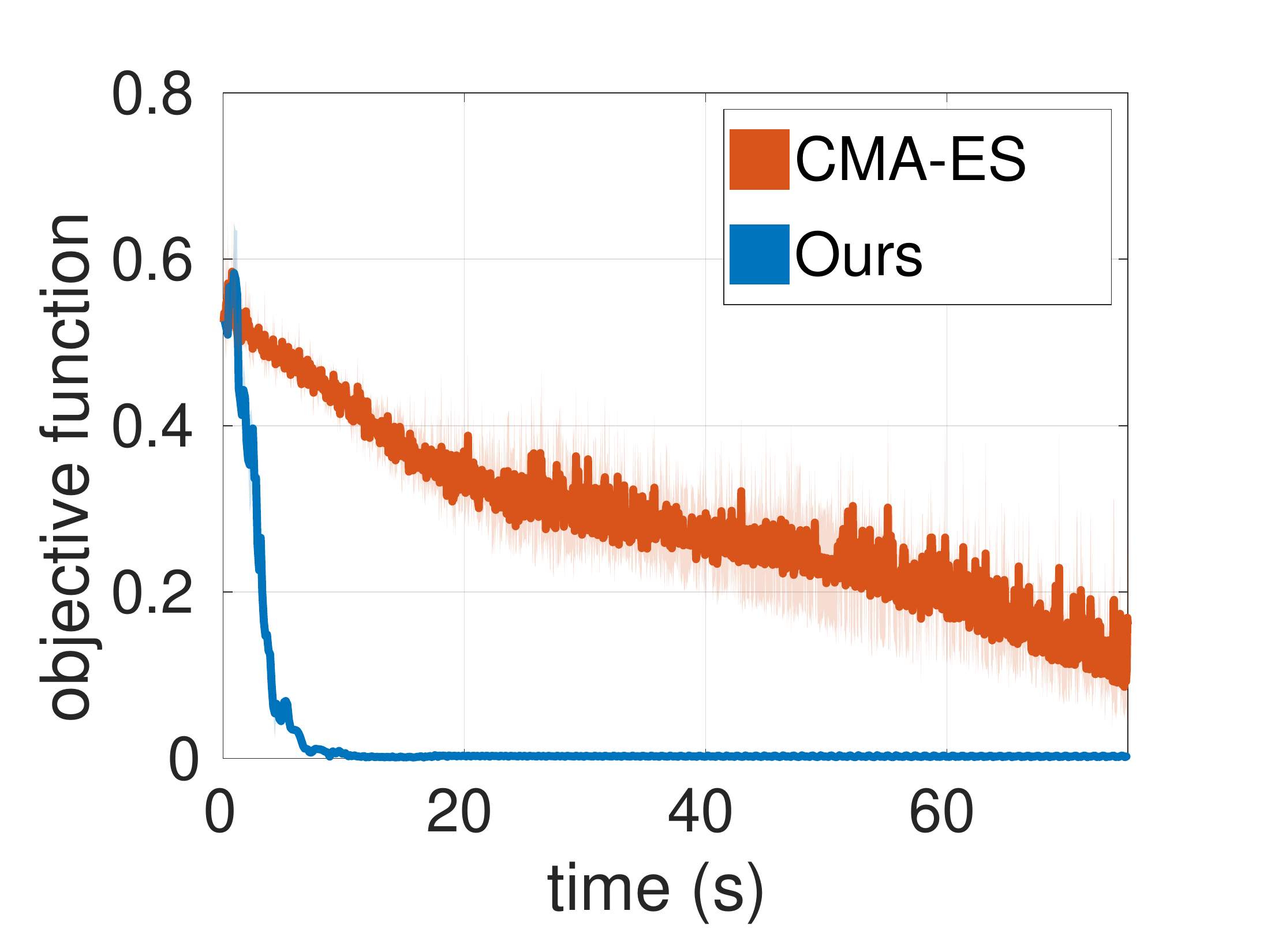} \\
    (a) Throwing & (b) Hitting
\end{tabular}
\vspace{-2mm}
\caption{{\bf Motion control.} The task is to optimize the torque vectors to (a) throw a ball with an articulated 9-DoF robot arm and (b) hit a golf ball with an articulated linkage. In both scenarios, the goal is for the ball to settle at a specified target location. SGD with gradients provided by our method converges within 20 and 50 steps, respectively, while gradient-free optimization fails to reach comparable accuracy even with more than an order of magnitude higher number of iterations. The third row indicates that even though CMA-ES does not compute gradients, our method is still faster overall in wall-clock time.}
\label{fig:inverse}
\end{figure}

\subsection{Applications}

\mypara{Motion control.}
Differentiable physics enables application of gradient-based optimization to learning and control problems that involve physical systems. In Figure~\ref{fig:inverse}, we show two physical tasks: (a) throwing a ball to a target location using a 9-DoF robot arm and (b) hitting a golf ball to a target location using an articulated linkage. These tasks are also shown in the supplementary video.

For both scenarios, a torque vector is applied to the joints in each time step. Assume there are $n$ time steps and $k$ DoFs. The torque variable to be optimized has $n\times k$ dimensions, and the objective function is the L2 distance from the ball's actual position to the target. The joint positions and the torques are initialized as $\bZero$. If no gradients are available, a derivative-free optimization algorithm such as CMA-ES~\cite{HansenTutorial} can be used. We plot the loss curves of our method and CMA-ES in Figure~\ref{fig:inverse}. 

In (a),
SGD with gradients from our simulator converges in 20 steps. In contrast, CMA-ES does not reach the same accuracy even after 500 steps. In (b),
SGD with gradients from our simulator converges in 50 steps, while CMA-ES does not reach the same accuracy even after 1500 steps.

\begin{figure}
\centering
\begin{tabular}{@{}c@{\hspace{0.2mm}}c@{}}
    \includegraphics[width=.45\linewidth]{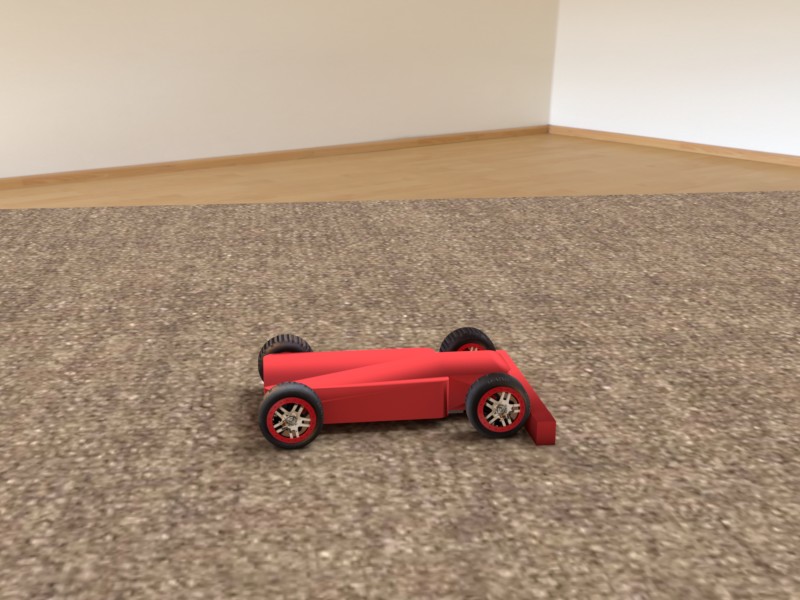} &
    \includegraphics[width=.47\linewidth]{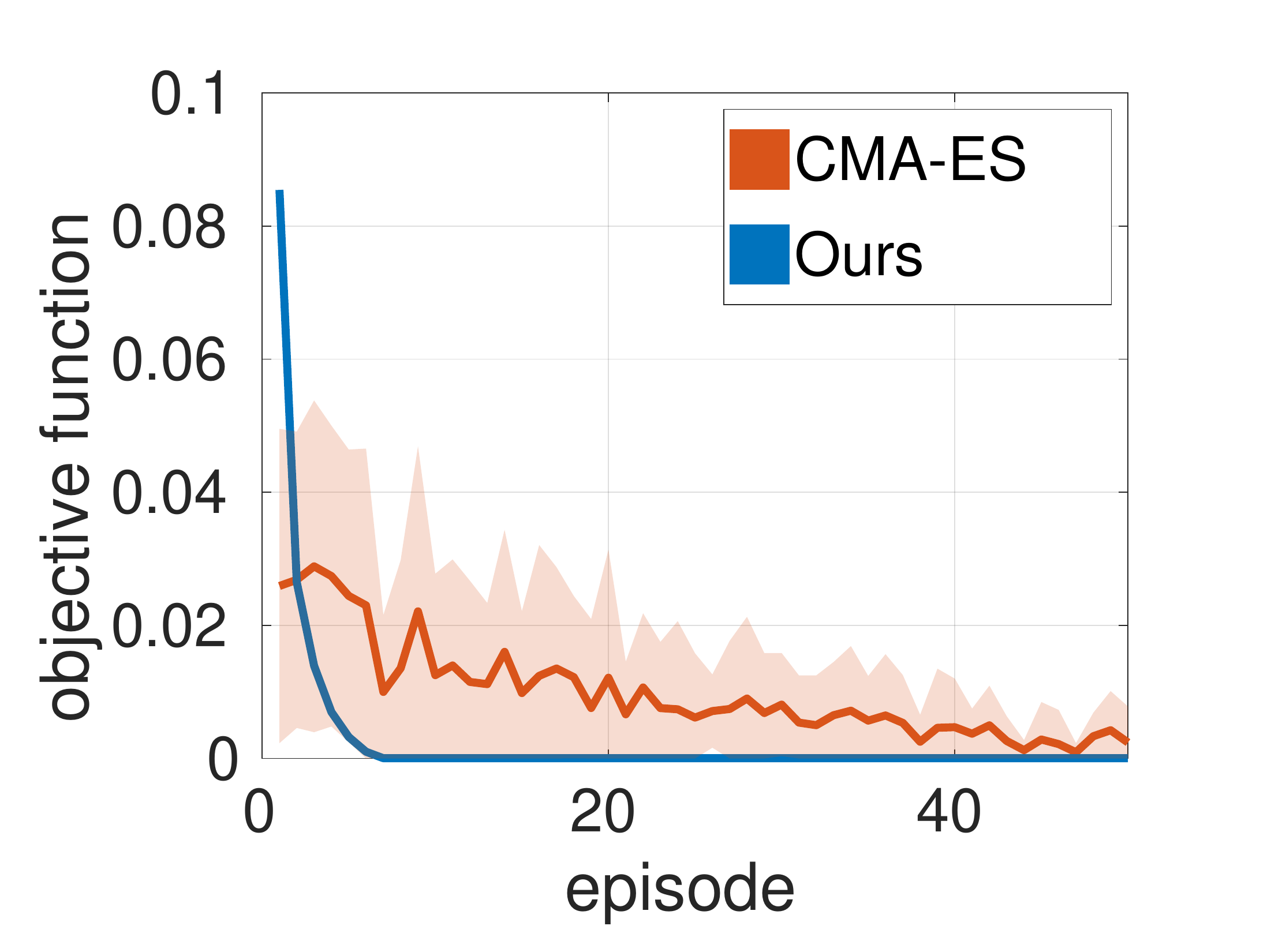} \\
    \small (a) A racecar  & \small (b) Loss per episode \\
    \includegraphics[width=.47\linewidth]{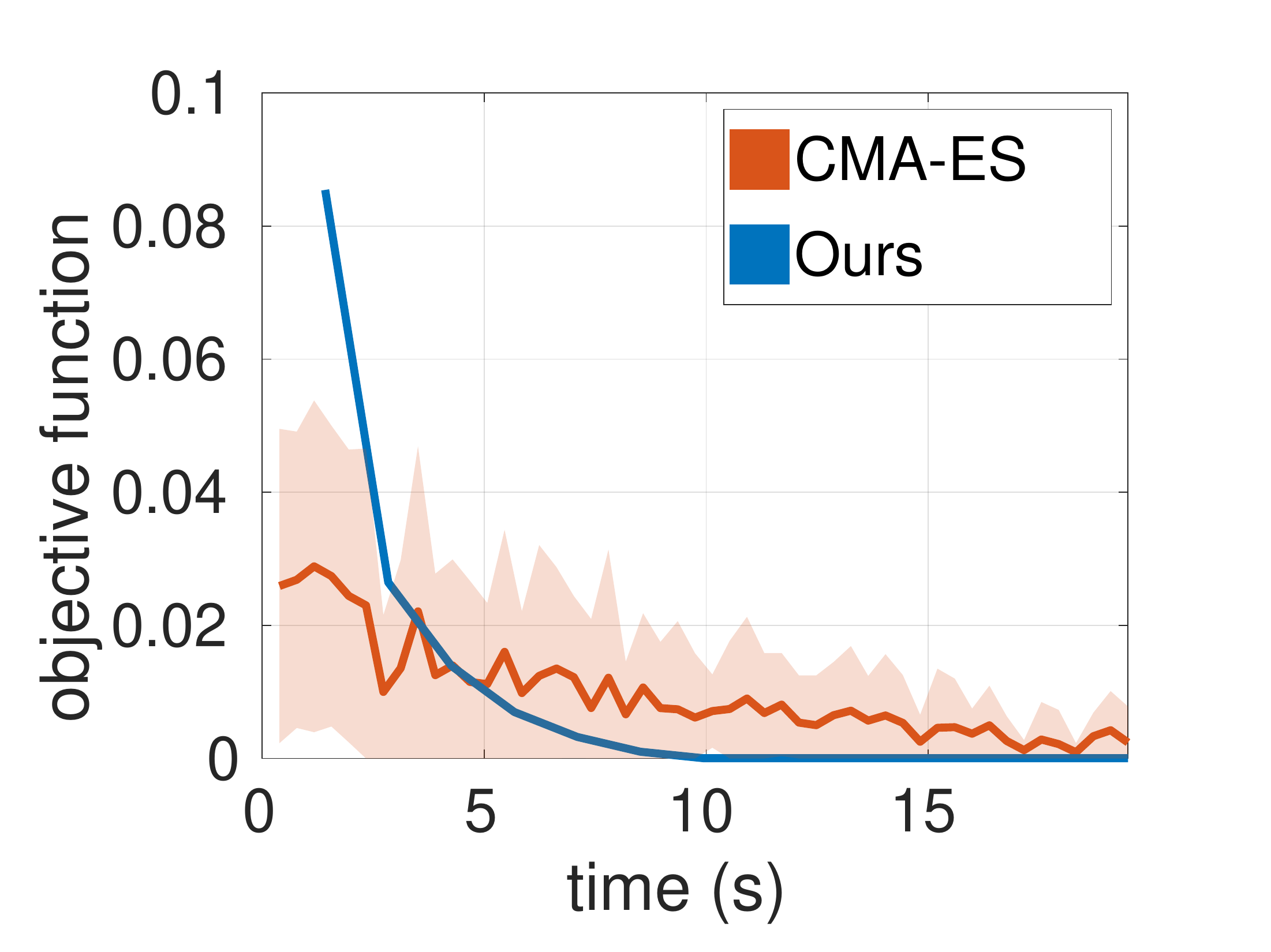} &
    \includegraphics[width=.47\linewidth]{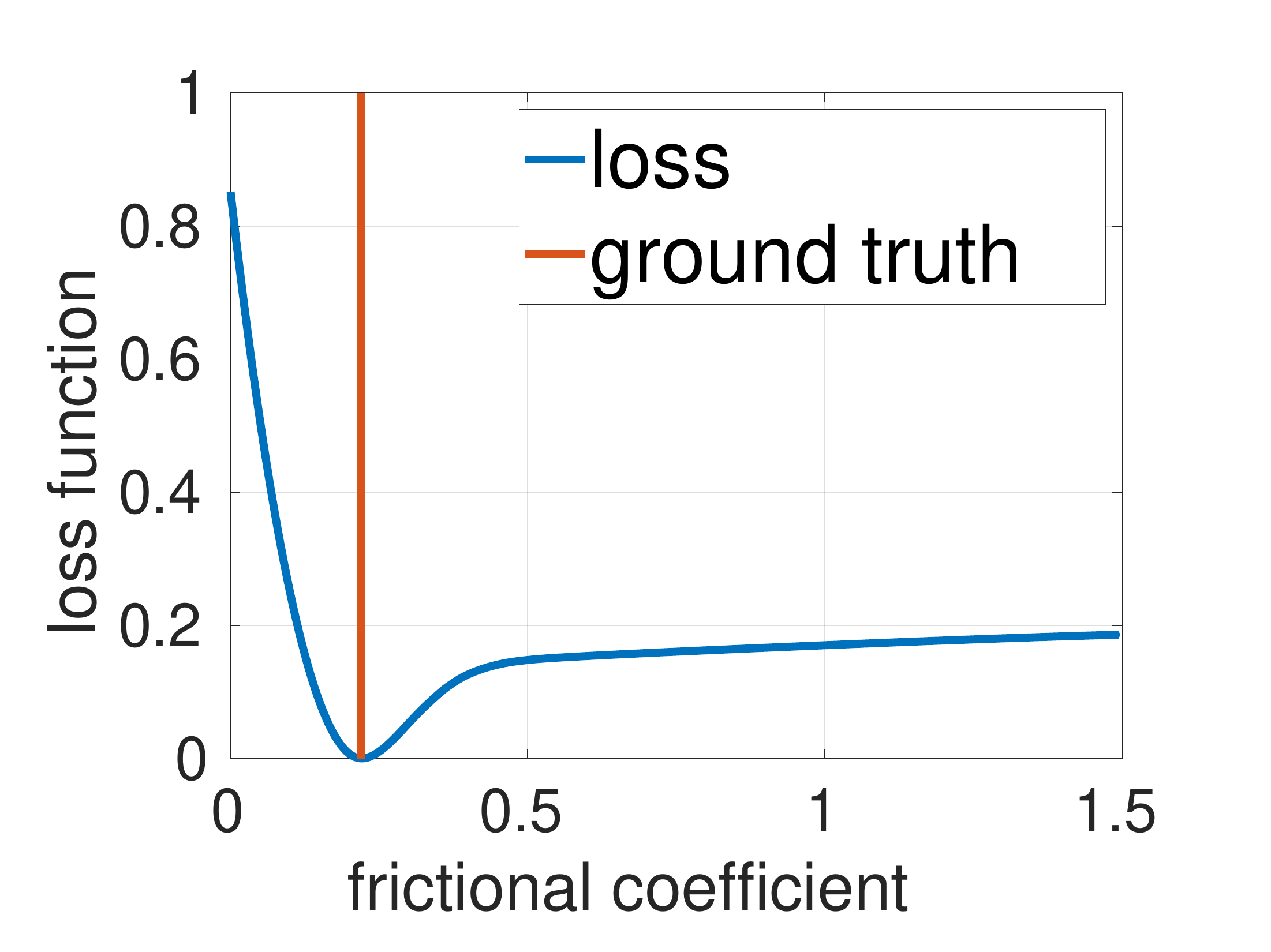}
    \\
    \small (c) Loss over time & \small (d) Loss landscape
\end{tabular}
\vspace{-2mm}
\caption{{\bf Parameter estimation.} The goal is to estimate the sliding friction coefficient that makes the racecar decelerate to a target location. (b,c) Loss curves in episode and wall-clock time, respectively. (d) Loss landscape plot. SGD with gradients computed by our method solves the task within 10 iterations, more accurately and faster than the gradient-free baseline.}
\vspace{-1em}
\label{fig:estimate}
\end{figure}

\mypara{Parameter estimation.}
Our method can estimate unknown physical parameters from observation. In Figure~\ref{fig:estimate}, a racecar starts with horizontal velocity $1 m/s$. The wheels and the steering system are articulated.
We estimate the sliding friction coefficient $\mu$ between the wheels and the ground such that the car stops at $x=0.8m$ at the end of the episode. At the initial guess $\mu=0.002$, the car reaches $x=1.0m$. We use gradient descent to optimize the estimated friction coefficient. After 10 iterations of SGD with gradients provided by our method, the car reaches the goal with $\mu=0.21$. In comparison, the gradient-free baseline takes multiple times longer to reach comparable objective values.



\section{Conclusion}
\label{sec:conclusion}
We have developed a differentiable simulator for articulated body dynamics that runs 10x faster with 100x smaller memory footprint in comparison to existing autodiff tools. To achieve this performance gain, we analyze the workload of each simulation step to better manage computation and memory. The adjoint method is used to compute the gradients of the simulation. We derive the adjoints of spatial algebra and the Gauss-Seidel solver. We then adapt the checkpointing method to deal with the sequential nature of articulated body simulation, reducing memory consumption by two orders of magnitude.

As a supporting contribution, we have explored two applications of differentiable physics to reinforcement learning with physical systems, reporting preliminary results that indicate that differentiable physics can accelerate learning. We have also presented application scenarios that clearly demonstrate the effectiveness of gradient-based optimization in motion control and parameter estimation with articulated physical systems.


\mypara{Acknowledgements.} This research is supported in part by Intel Corporation, National Science Foundation, Dr. Barry Mersky and Capital One Endowed Professorship, and Elizabeth Stevinson Iribe Endowed Chair Professorship.

{
	\bibliographystyle{icml2021}
	\bibliography{paper}
}

\clearpage

\section*{\LARGE Supplementary Information}

\begin{appendices}
\label{sec:appendix}

\section{Discussion of RL Applications}
\label{sec:DiscussRL}

For the policy enhancement method used in $n$-link pendulum,
we use off-policy training and weigh the relevance of each sample according to the distance to the current action:
\begin{gather}
    w=\exp(-\|\aa-\aa_0\|)
\end{gather}
where $\aa$ is the policy network output, and $\aa_0$ is the action sampled from the history pool.
For every epoch, we sample from the environment 1,000 times and train the model after each sample.
We train both models from the simplest 1-link to the most difficult 7-link setting, each experiment repeated 10 times to obtain stable statistics.
In Table~\ref{tab:pendulum}, we report the statistics of the maximum return for pendulums of different numbers of links.
As mentioned in the main text, the performance of MBPO degrades while ours does not.
The reason behind it is that, during the policy update, our method does not need to fully depend on the critic function.
By making use of the gradients as shown in Eq.~\ref{equ:fakeloss}, the policy network can first optimize the short-term reward ($\partial r/\partial a$) before the critic is up to date.
When the critic network is able to reveal the true value of the state-action pair, the policy can then converge as well to the long term goal.
On the other hand, traditional RL methods heavily depend on the convergence of the critic network, thus having a hard time searching for the best action in a large space, especially when the links are attached sequentially.

In Fig.~\ref{fig:ant} that shows our MuJoCo Ant experiment results, we observe an even faster convergence than MBPO, while MBPO already outperforms most of the RL algorithms.
This is because by generating nearby samples, the feedback from the environment is no longer a single point but a patch as an approximation of the shape of the state transitioning function.
It can greatly help the critic function learn the correct estimation of the values not only for the observation point but also nearby ones, thereby achieving faster convergence on the value estimation and eventually the policy model.

\noindent
{\bf Limitations and future directions.} We also observe limitations on these two proposed techniques.
We observed that the sample enhancement does not provide better overall rewards in the `pendulum' test, and the policy enhancement does not provide faster convergence in the `ant' test.
We attribute this to a number of possible causes.
Sample enhancement enriches the interaction history with the environment, but it is inherently limited because extra samples can only be generated near the true sample, it may not help the agent jump out of the local minima.
Policy enhancement relies heavily on the assumption that the actions drawn from the training batch are the same as or similar to the action that the policy takes at the current moment: it is considered as an on-policy learning algorithm.
Therefore, it faces the same limitation, which is poor sample efficiency, as other on-policy algorithms as well.
These two techniques cannot be combined together because the generated samples from the sample enhancement do not have accurate first-order gradients at their location, as mentioned in the paper.

While we have shown the two techniques to be a proof-of-concept demonstration that differentiable physics can help improve performance of reinforcement learning using an appropriate method depending on different scenarios, there remain several directions to further explore.  We hope that our preliminary investigation on differentiable physics integrated with RL, as presented in this paper, will stimulate more study on the coupling of differentiable physics with reinforcement learning.

\begin{table*}[!htb]
\centering
\ra{1.1}
 \resizebox{1\linewidth}{!}{
	\small
	\begin{tabular}{l|ccccccc}
		\toprule
		\# of links & 1 & 2  &  3 & 4 & 5 & 6 & 7   \\
		\hline
		Ours    &
\textbf{0.49$\pm$	0.01}&
\textbf{1.93$\pm$	0.08}&
\textbf{4.28$\pm$	0.25}&
\textbf{7.81$\pm$	0.19}&
\textbf{12.31$\pm$	0.16}&
\textbf{17.78$\pm$	0.24}&
\textbf{24.34$\pm$	0.11}\\
MBPO&
\textbf{0.49$\pm$	0.01}&
1.90$\pm$	0.02&
\textbf{4.28$\pm$	0.22}&
7.50$\pm$	0.57&
11.13$\pm$	0.99&
15.02$\pm$	1.45&
19.97$\pm$	2.33\\
		Maximum &0.5&2&4.5&8&12.5&18&24.5\\
		\bottomrule
	\end{tabular}
 }
\vspace{-2mm}
\caption{Maximum reward on $n$-link pendulum. Our method has higher reward scores than MBPO, as the system complexity increases.}
\label{tab:pendulum}
\end{table*}

\section{Ablation Study for Checkpointing Scheme}
In this section, we conduct an ablation study to show why we choose to save the checkpoints each time step, 
instead of having a lower frequency for checkpointing. We vary the intervals between checkpoints and profile the peak memory usage and backpropagation time per step. Our default design is interval=1 such that checkpoints are stored every step.

If interval=$k$, our method needs to compute the gradients $\overline{x_i},\overline{x_{i+1}},..,\overline{x_{i+k-1}}$ given the gradients $\overline{x_{i+k}}$ and the checkpoint $x_i$. The intermediate variables in step $i+k-1, i+k-2,...,i$ have to be resumed from $x_i$, taking $k, i-1,...,1$ steps of forward simulations, respectively. Therefore, the total forward steps used to resume variables here are $(k+1)k/2$. If there are $n$ simulation steps in total, the average time of backpropagation per step should be 
\begin{equation}
    (k+1)/2\cdot t_{forawrd} +t_{backward},
\end{equation}
where $t_{simulation},t_{adjoint}$ are the time for one step of simulation and adjoint method, respectively. On the other hand, the peak memory usage is 
\begin{equation}
    n/k\cdot m_{checkpoint}+m_{simulation},
\end{equation}
where $m_{checkpoint}$ is the size of one checkpoint; $m_{simulation}$ is the memory usage of one step, which is the size of all intermediate variables. Since $m_{simulation} \ggreater m_{checkpoint}$, the increase of $k$ would not significantly reduce the memory usage but will slow down the backpropagation linearly.

\begin{figure}
\centering
\begin{tabular}{@{}c@{\hspace{1mm}}c@{\hspace{1mm}}c@{}}
    \includegraphics[width=.48\linewidth]{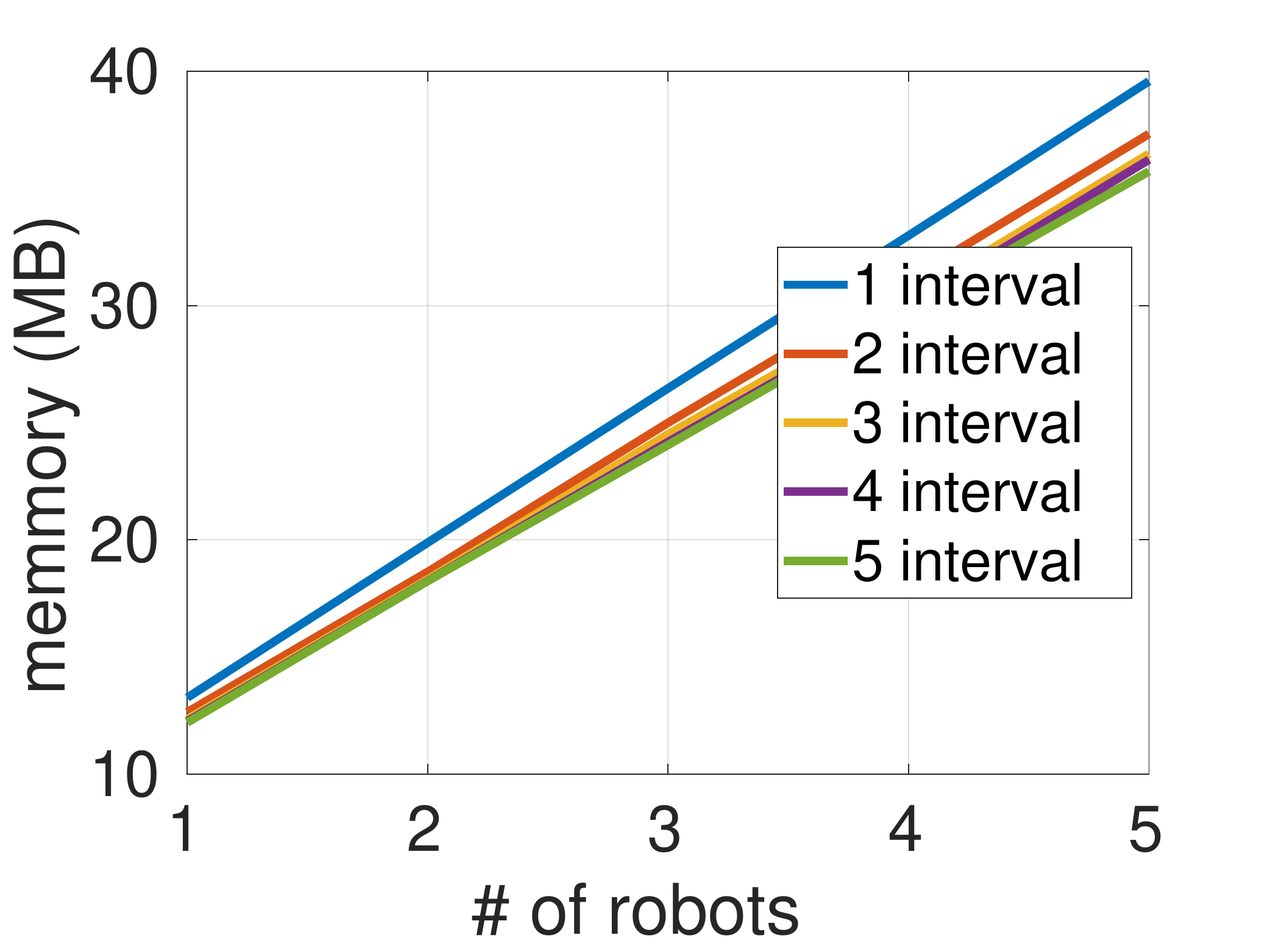} &
    \includegraphics[width=.48\linewidth]{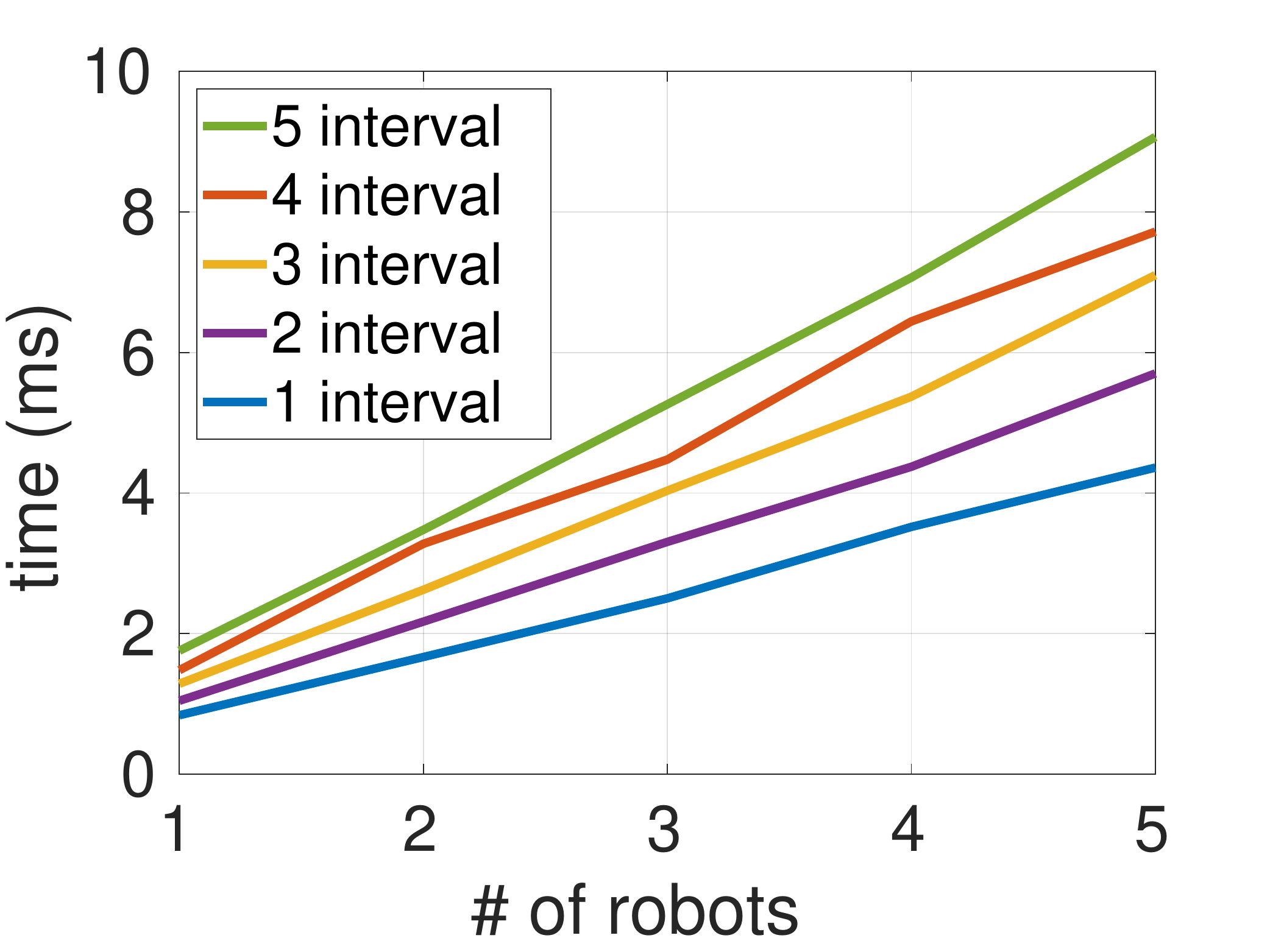} \\
    (a) Memory  & (b) Time
\end{tabular}
\vspace{-2mm}
\caption{Ablation study for the checkpointing scheme.}
\vspace{-1em}
\label{fig:ab_ckpt}
\end{figure}

In Figure~\ref{fig:ab_ckpt}, the experiments have similar setup as in Section~\ref{sec:autodiff} where we simulate 10 Laikago robots for 5,000 steps. The memory and the time scale linearly w.r.t. the number of the robots. When the checkpoint intervals get larger, the memory usage decreases gradually to the lower bound of $m_{simulation}$, but the average backpropagation time increases linearly. Since the memory usage is low enough even for interval $k=1$, we choose to checkpoint every step to have a faster speed and set the interval $k=1$.

\section{The Adjoints of Contact Solver}
In the collision solver, we construct a Mixed Linear Complementarity Problem (MLCP)~\cite{Stepien2013}:
\begin{equation}
\begin{split}
    &\aa = \AA\xx+\bb\  \\
    \textrm{ s.t. \hspace{2mm} } \bm{0}\leq &\aa \perp \xx\geq \bm{0}\textrm{\hspace{1mm} and \hspace{1mm}} \cc_- \leq \xx  \leq\cc_+ ,
\end{split}
\label{eq:mlcp}
\end{equation}
where $\xx$ is the new collision-free state, $\AA$ is the inertial matrix, $\bb$ contains the relative velocities, and $\cc_-, \cc_+$ are the lower bound and upper bound constraints, respectively. In the forward pass, $\AA,\bb$ are known variables, and the projected Gauss-Seidel (PGS) method is used to solve for $\xx$. In Algorithm~\ref{alg:adjointpgs}, we are computing the adjoint of $\AA,\bb$ given the adjoint $\overline{\xx}$. For simplicity, we assume there are no constraints $\cc_-, \cc_+$ and it is now a pure Gauss-Seidel method which solves $\AA\xx+\bb=\mathbf{0}$. Basically, we store variables in $\VV_d,\VV_{x1}$, and $\VV_{x2}$. In the backward pass, we load the variables and compute the adjoints in a reverse order.
\begin{algorithm}[!htb]
\caption{Adjoint of Gaussian-Siedel Solver}
\label{alg:adjointpgs}
\begin{algorithmic}
\STATE {\bfseries Input:} matrix $\AA$, vector$\bb$, adjoint $\overline{\xx}$, iterations $niter$
\STATE {\bfseries Output:} adjoints $\overline{\AA}$, $\overline{\bb}$ 
\STATE Initialize vectors $\VV_d(niter,rows(\AA))$, 
\STATE $\VV_{x1}(niter,rows(\AA))$, $\VV_{x2}(niter,rows(\AA))$.
\STATE \# forward pass
\FOR{$t=1$ {\bfseries to} $niter$}
\STATE $\VV_{x1}(t)=\xx$
\FOR{$i=1$ {\bfseries to} $rows(\AA)$}
\STATE $d = 0$
\FOR{$j=1$ {\bfseries to} $rows(\AA)$}
\IF{$j\neq i$}
\STATE $d=d+\AA(i,i)*\xx(j)$
\ENDIF
\ENDFOR
\STATE $\VV_d(t,i)=d$
\STATE $\xx(i)=(\bb(i)-d) / \AA(i,i)$
\ENDFOR
\STATE $\VV_{x2}(t,i)=\xx(i)$
\ENDFOR
\STATE \# start backward process
\STATE $\AA = \mathbf{0}$, $\bb=\mathbf{0}$
\FOR{$t$niter {\bfseries to} 1}
\FOR{$i=rows(\AA)$ {\bfseries to} 1}
\STATE $\overline{d}=0$
\STATE $\overline{d}=\overline{d}-\overline{\xx}(i)/\AA(i,i)$
\STATE $\overline{\bb}=\overline{\bb}+\overline{\xx}(i)/\AA(i,i)$
\STATE $\overline{\AA}(i,i)=\overline{\AA}(i,i)-\overline{x}(i)*[\bb(i)-\VV_d(t,i)]/\AA(i,i)^2$
\STATE $\overline{\xx}(i)=0$
\FOR{$j=rows(\AA)$ {\bfseries to} i+1}
\STATE $\overline{\AA}(i,j)=\overline{\AA}(i,j)d+\overline{d}*\VV_{x1}(t,j)$
\STATE $\overline{\xx}(j)=\overline{\xx}(j)+\overline{d}*\AA(i,j)$
\ENDFOR
\FOR{$j=i-1$ {\bfseries to} 1}
\STATE $\overline{\AA}(i,j)=\overline{\AA}(i,j)d+\overline{d}*\VV_{x2}(t,j)$
\STATE $\overline{\xx}(j)=\overline{\xx}(j)+\overline{d}*\AA(i,j)$
\ENDFOR
\ENDFOR
\ENDFOR
\end{algorithmic}
\end{algorithm}

\section{Model Mismatch/Transplant}
Our articulated body simulation has similar functionality to MuJoCo~\cite{todorov2012mujoco} in terms of physical and contact modeling.
We first trained the `MuJoCo Ant' model using our simulation as the environment.
Upon convergence, we directly tested our model in MuJoCo, while keeping all robot configuration the same.
As shown in Table~\ref{tab:modeltransfer}, without any retraining, the trained model is able to obtain non-trivial reward in MuJoCo.
\begin{table}[!htb]
\centering
\ra{1.1}
	\small
	\begin{tabular}{l|c}
		\toprule
		Method & Reward in MuJoCo   \\
		\hline
w/o pretraining	&868$\pm$77\\
w/ pretraining (Ours)	&\textbf{1735$\pm$1066}\\
		\bottomrule
	\end{tabular}
\vspace{-2mm}
\caption{Model transplant results on MuJoCo Ant. After pretraining in our simulation environment, the model can be directly transplanted to a new environment, MuJoCo, with non-trivial reward. This shows that our simulator behaves similarly with MuJoCo.}
\label{tab:modeltransfer}
\end{table}

\section{Other Limitations and Future Directions}
In addition to some possible directions to further explore, as mentioned in Appendix~\ref{sec:DiscussRL}, our method can benefit from additional enhancement. 
The current simulator can potentially provide more options for users. 
For example, we have not provided options to select different contact models. The current LCP formula using iterative solver is fast but sometimes not accurate enough. Implementing extra contact models like exact LCP solvers or convex soft contact model can help the simulator adapt to more applications.
Moreover, we assume that each link is a rigid body. It would be natural to extend this work to support deformable objects such as cloth~\cite{todorov2012mujoco,Qiao2020Scalable}. 

We also plan to build on top of the current simulator to provide more comprehensive tool sets for users. Furthermore, it is worth exploring how differentiable physics can improve the sampling strategies and optimization of deep reinforcement learning models.  Lastly, we hope to deploy our algorithms on real-world robots to perform a wide range of complex tasks.

\section{Code for Profiling Autodiff Tools}
\mypara{ADF.} We post below the code to run the backpropagation of ADF~\cite{Leal2018Autodiff}. As we described in Section~\ref{sec:autodiff}, ADF can be used to run the forward simulation of articulated body dynamics but fails to complete the backpropagation in a reasonable time. We found out that ADF will greatly slow down when the computation graph goes deeper. Here is an example where we repeat a simple operation iteratively. When the iteration number is 28, it already takes a long time to backpropagate. In articulated body simulations, the depth of computation is far larger than 28, so ADF fails to compute the corresponding gradients.
\begin{lstlisting}
#include <autodiff/reverse.hpp>
using namespace autodiff;

// The single-variable function for 
// which derivatives are needed
var f(var x)
{
    int num_iter = 28;  
    for (int i=0; i<num_iter; i++)
      x = x + 1 / x;
    return x;
}

int main()
{
    var x = 2.0;   // the input variable x
    var u = f(x);  // the output variable u

    // evaluate the derivative of u 
    // with respect to x
    auto [ux] = derivatives(u, wrt(x)); 

}
\end{lstlisting}

\mypara{JAX.} JAX is based on Python and is hard to integrate into our C++ simulator. Therefore, we write simplified Python code to do the profiling. The input parameters to this code are the number of robots and simulation length. Each robot in the simulation environment of Section~\ref{sec:autodiff} has 37 DoFs, so we multiply 37 by the number of robots and get the dimension of the states variable. The operations in one simulation loop are far simpler than the articulated body dynamics. 

\begin{lstlisting}
from jax import grad
import jax.numpy as jnp
from jax import jit
import numpy as np
import sys 

class RunBW(object):
  def __init__(self, dim, n_step):
    self.dim = dim
    self.n_step = n_step
    
  def test(self, x):  
    dim = self.dim
    n_step = self.n_step
    for s in range(n_step):
      var = []
      for k in range(dim):
        var.append(x[k])
      for i in range(dim):
        k = i % (dim -1) + 1
        var[k] = var[k-1] + 1/var[k]
      var = [jnp.expand_dims(v, 0) for 
        v in var]
      x = jnp.concatenate(var, 0)
    return jnp.sum(x)

if __name__ == '__main__':
  dim = 37 * int(sys.argv[1])
  n_step = int(sys.argv[2])
  rb = RunBW(dim, n_step)
  ini_state = np.ones([dim])
  grad_test = grad(rb.test)  
  result = grad_test(ini_state)
  result = rb.test(ini_state)
\end{lstlisting}

\onecolumn

\section{Differentiation of Articulated Body Algorithm}
This section describes the list of adjoint operations. We start with basic operators and go over all adjoints used in our implementation. $\overline{(\cdot)}$ is denoted as the adjoint of a variable.\\
Scalar multiply:
\begin{equation}
\begin{split}
    a &= bc \\
    \PD{\phi}{b} &=\PD{\phi}{a}\PD{a}{b} \\
    \adj{b}{} &= \adj{a}{} c,\adj{c}{} = b\adj{a}{} 
\end{split}
\end{equation}

Matrix vector multiply:
\begin{equation}
\begin{split}
    \vv&=\mm\aa \\
    \adj{\mm}{ij}&=\adj{\vv}{i}\aa_j,\adj{\aa}{j}=\sum \adj{\vv}{k}\mm_{kj} \\
    \adj{\mm}{}&=\adj{\vv}{}\aa^T\\
    \adj{\aa}{}&=\mm^T\adj{\vv}{}
\end{split}
\end{equation}

Matrix multiply:
\begin{equation}
\begin{split}
    \cc&=\aa\bb \\
    \adj{\aa}{}&=\adj{\cc}{}\bb^T \\
    \adj{\bb}{}&=\aa^T\adj{\cc}{}
\end{split}
\end{equation}

Matrix $\AA^T\BB\AA$:
\begin{equation}
\begin{split}
    \cc&=\aa^T\bb\aa \\
    \adj{\aa}{}&=\bb^T\aa\adj{\cc}{}+\bb\aa\adj{\cc}{}^T \\
    \adj{\bb}{}&=\aa\adj{\cc}{}\aa^T
\end{split}
\end{equation}


Matrix inverse:
\begin{equation}
\begin{split}
    \bb&=\aa^{-1} \\
    \PD{\bb}{\qq}&=-\bb\PD{\aa}{\qq}\bb=-\sum_k\sum_t \bb_{ij}\PD{\aa_{jk}}{\qq}\bb_{kt} \\
    \adj{\aa}{jk}&=-\sum_i\sum_t\bb_{ij}\bb_{kt}\adj{\bb}{it} \\
    \adj{\aa}{}&=-\bb^T\adj{\bb}{}\bb^T
\end{split}
\end{equation}

Multiply three matrices:
\begin{equation}
\begin{split}
    \mm&=\aa\bb\cc \\
    \adj{\aa}{}&=\adj{\mm}{}\cc^T\bb^T\\
    \adj{\bb}{}&=\aa^T\adj{\mm}{}\cc^T\\
    \adj{\cc}{}&=\bb^T\aa^T\adj{\mm}{}\\
\end{split}
\end{equation}

Vector cross product:
\begin{equation}
\begin{split}
    \vv=\aa\times\bb&=[\aa_2\bb_3-\aa_3\bb_2,\aa_3\bb_1-\aa_1\bb_3,\aa_1\bb_2-\aa_2\bb_1] \\
    \adj{\aa}{}&=[-\adj{\vv}{2}\bb_3+\adj{\vv}{3}\bb_2,\adj{\vv}{1}\bb_3-\adj{\vv}{3}\bb_1,-\adj{\vv}{1}\bb_2+\adj{\vv}{2}\bb_1]=-\adj{\vv}{}\times\bb\\
    \adj{\bb}{}&=[\adj{\vv}{2}\aa_3-\adj{\vv}{3}\aa_2,-\adj{\vv}{1}\aa_3+\adj{\vv}{3}\aa_1,\adj{\vv}{1}\aa_2-\adj{\vv}{2}\aa_1]=\adj{\vv}{}\times\aa\\
\end{split}
\end{equation}

Vector norm:
\begin{equation}
\begin{split}
    a&=\sqrt{\bb\cdot\bb}\\
    \PD{a}{\qq}&=-\frac{1}{2}\frac{1}{\sqrt{\bb\cdot\bb}}(\PD{\bb}{\qq}\cdot\bb+\bb\cdot\PD{\bb}{\qq})=\frac{1}{a}\PD{\bb}{\qq}\cdot\bb\\
    \adj{\bb}{}&=\frac{\adj{a}{}}{a}\bb
\end{split}
\end{equation}

Spatial transform: apply():
\begin{equation}
\left[
\begin{array}{c}
\bb_1\\
\bb_2
\end{array}\right]=
st_{apply}(
\left[
\begin{array}{c}
\EE\\
\rr
\end{array}\right],
\left[
\begin{array}{c}
\aa_1 \\
\aa_2
\end{array}\right])=
\left[
\begin{array}{c}
\EE\aa_1 \\
\EE(\aa_2-\rr\times\aa_1 )
\end{array}\right]
\end{equation}\label{eq:st-apply}

\begin{equation}
\left[
\begin{array}{c}
\adj{\EE}{}\\
\adj{\rr}{}
\end{array}\right]=
\left[
\begin{array}{c}
\aa_1\adj{\bb}{1}^T+(\aa_2-\rr\times\aa_1)\adj{\bb}{2}^T\\
\EE^T\adj{\bb}{2}\times\aa_1
\end{array}\right]
,
\left[
\begin{array}{c}
\adj{\aa}{1} \\
\adj{\aa}{2}
\end{array}\right]=
\left[
\begin{array}{c}
\EE^T\adj{\bb}{1}-\EE^T\adj{\bb}{2}\times\rr \\
\EE^T\adj{\bb}{2}
\end{array}\right]
\end{equation}\label{eq:adj-st-apply}

Spatial transform: apply-inv():
\begin{equation}
\left[
\begin{array}{c}
\bb_1\\
\bb_2
\end{array}\right]=
st_{apply-inv}(
\left[
\begin{array}{c}
\EE\\
\rr
\end{array}\right],
\left[
\begin{array}{c}
\aa_1 \\
\aa_2
\end{array}\right])=
\left[
\begin{array}{c}
\EE^T\aa_1 \\
\EE^T\aa_2+\rr\times(\EE^T\aa_1 )
\end{array}\right]
\end{equation}\label{eq:st-apply-inv}

\begin{equation}
\left[
\begin{array}{c}
\adj{\EE}{}\\
\adj{\rr}{}
\end{array}\right]=
\left[
\begin{array}{c}
\aa_1(\adj{\bb}{1}+\adj{\bb}{2}\times \rr)^T+\aa_2\adj{\bb}{2}^T\\
-\adj{\bb}{2}\times\EE^T\aa_1
\end{array}\right]
,
\left[
\begin{array}{c}
\adj{\aa}{1} \\
\adj{\aa}{2}
\end{array}\right]=
\left[ 
\begin{array}{c}
\EE(\adj{\bb}{1}+\adj{\bb}{2}\times\rr) \\
\EE\adj{\bb}{2}
\end{array}\right]
\end{equation}\label{eq:adj-st-inv}

Spatial transform: apply-trans():
\begin{equation}
\left[
\begin{array}{c}
\bb_1\\
\bb_2
\end{array}\right]=
st_{apply-trans}(
\left[
\begin{array}{c}
\EE\\
\rr
\end{array}\right],
\left[
\begin{array}{c}
\aa_1 \\
\aa_2
\end{array}\right])=
\left[
\begin{array}{c}
\EE^T\aa_1+\rr\times (\EE^T\aa_2) \\
\EE^T\aa_2
\end{array}\right]
\end{equation}\label{eq:st-apply-trans}

\begin{equation}
\left[
\begin{array}{c}
\adj{\EE}{}\\
\adj{\rr}{}
\end{array}\right]=
\left[
\begin{array}{c}
\adj{\bb}{1}\aa_1^T+(\adj{\bb}{1}\times\rr+\adj{\bb}{2})\aa_2^T \\
-\adj{\bb}{1}\times\EE^T\aa_2
\end{array}\right]
,
\left[
\begin{array}{c}
\adj{\aa}{1} \\
\adj{\aa}{2}
\end{array}\right]=
\left[
\begin{array}{c}
\EE\adj{\bb}{1} \\
\EE(\adj{\bb}{1}\times\rr+\adj{\bb}{2})
\end{array}\right]
\end{equation}\label{eq:adj-st-trans}

Spatial transform: apply-invtrans():
\begin{equation}
\left[
\begin{array}{c}
\bb_1\\
\bb_2
\end{array}\right]=
st_{apply-invtrans}(
\left[
\begin{array}{c}
\EE\\
\rr
\end{array}\right],
\left[
\begin{array}{c}
\aa_1 \\
\aa_2
\end{array}\right])=
\left[
\begin{array}{c}
\EE^T(\aa_1-\rr\times\aa_2) \\
\EE^T\aa_2
\end{array}\right]
\end{equation}\label{eq:st-apply-invtrans}

\begin{equation}
\left[
\begin{array}{c}
\adj{\EE}{}\\
\adj{\rr}{}
\end{array}\right]=
\left[
\begin{array}{c}
(\aa_1-\rr\times\aa_2)\adj{\bb}{1}^T+\aa_2\adj{\bb}{2}^T \\
\EE\adj{\bb}{1}\times\aa_2
\end{array}\right]
,
\left[
\begin{array}{c}
\adj{\aa}{1} \\
\adj{\aa}{2}
\end{array}\right]=
\left[
\begin{array}{c}
\EE\adj{\bb}{1} \\
-\EE\adj{\bb}{1}\times\rr+\EE\adj{\bb}{2}
\end{array}\right]
\end{equation}\label{eq:adj-st-invtrans}

Spatial transform: multiply():
\begin{equation}
\left[
\begin{array}{c}
\EE_0\\
\rr_0
\end{array}\right]=
st_{multiply}(
\left[
\begin{array}{c}
\EE_1\\
\rr_1
\end{array}\right],
\left[
\begin{array}{c}
\EE_2\\
\rr_2
\end{array}\right])=
\left[
\begin{array}{c}
\EE_1 \EE_2\\
\rr_2+\EE_2^T\rr_1
\end{array}\right]
\end{equation}\label{eq:st-multiply}

\begin{equation}
\left[
\begin{array}{c}
\adj{\EE}{1}\\
\adj{\rr}{1}
\end{array}\right]=
\left[
\begin{array}{c}
\adj{\EE}{0}\EE_2^T \\
\EE_2\adj{\rr}{0}
\end{array}\right]
,
\left[
\begin{array}{c}
\adj{\EE}{2}\\
\adj{\rr}{2}
\end{array}\right]=
\left[
\begin{array}{c}
\EE_1^T\adj{\EE}{0}+\rr_1\adj{\rr}{0}^T \\
\adj{\rr}{0}
\end{array}\right]
\end{equation}\label{eq:adj-st-multiply}

Spatial motion: crossm():
\begin{equation}
\left[
\begin{array}{c}
\ww_0\\
\vv_0
\end{array}\right]=
sm_{crossm}(
\left[
\begin{array}{c}
\ww_1\\
\vv_1
\end{array}\right],
\left[
\begin{array}{c}
\ww_2\\
\vv_2
\end{array}\right])=
\left[
\begin{array}{c}
\ww_1 \times \ww_2\\
\ww_1\times\vv_2+\vv_1\times\ww_2
\end{array}\right]
\end{equation}\label{eq:sm-crossm}

\begin{equation}
\left[
\begin{array}{c}
\adj{\ww}{1}\\
\adj{\vv}{1}
\end{array}\right]=
\left[
\begin{array}{c}
-\adj{\ww}{0}\times\ww_2-\adj{\vv}{0}\times\vv_2 \\
-\adj{\vv}{0}\times\ww_2
\end{array}\right]
,
\left[
\begin{array}{c}
\adj{\ww}{2}\\
\adj{\vv}{2}
\end{array}\right]=
\left[
\begin{array}{c}
\adj{\ww}{0}\times\ww_1+\adj{\vv}{0}\times\vv_1 \\
\adj{\vv}{0}\times\ww_1
\end{array}\right]
\end{equation}\label{eq:adj-sm-crossm}

Spatial motion: crossf():
\begin{equation}
\left[
\begin{array}{c}
\ww_0\\
\vv_0
\end{array}\right]=
sm_{crossf}(
\left[
\begin{array}{c}
\ww_1\\
\vv_1
\end{array}\right],
\left[
\begin{array}{c}
\ww_2\\
\vv_2
\end{array}\right])=
\left[
\begin{array}{c}
\ww_1 \times \ww_2+\vv_1\times\vv_2\\
\ww_1\times\vv_2
\end{array}\right]
\end{equation}\label{eq:sm-crossf}

\begin{equation}
\left[
\begin{array}{c}
\adj{\ww}{1}\\
\adj{\vv}{1}
\end{array}\right]=
\left[
\begin{array}{c}
-\adj{\ww}{0}\times\ww_2-\adj{\vv}{0}\times\vv_2 \\
-\adj{\ww}{0}\times\vv_2
\end{array}\right]
,
\left[
\begin{array}{c}
\adj{\ww}{2}\\
\adj{\vv}{2}
\end{array}\right]=
\left[
\begin{array}{c}
\adj{\ww}{0}\times\ww_1\\
\adj{\ww}{0}\times\vv_1+\adj{\vv}{0}\times\ww_1
\end{array}\right]
\end{equation}\label{eq:adj-sm-crossf}

Spatial dyad: mul-ori():
\begin{equation}
\left[
\begin{array}{c}
\ww_0\\
\vv_0
\end{array}\right]=
sd_{mul-inv}(
\left[
\begin{array}{cc}
\mm_{11} & \mm_{12}\\
\mm_{21} & \mm_{22}
\end{array}\right],
\left[
\begin{array}{c}
\ww\\
\vv
\end{array}\right])=
\left[
\begin{array}{c}
\mm_{11}\ww+\mm_{12}\vv\\
\mm_{21}\ww+\mm_{22}\vv
\end{array}\right]
\end{equation}\label{ed:sd-mul-ori}

\begin{equation}
\left[
\begin{array}{cc}
\adj{\mm}{11}&\adj{\mm}{12}\\
\adj{\mm}{21}&\adj{\mm}{22}
\end{array}\right]=
\left[
\begin{array}{cc}
\adj{\ww}{0}\ww^T & \adj{\ww}{0}\vv^T \\
\adj{\vv}{0}\ww^T & \adj{\vv}{0}\vv^T
\end{array}\right]
,
\left[
\begin{array}{c}
\adj{\ww}{}\\
\adj{\vv}{}
\end{array}\right]=
\left[
\begin{array}{c}
\mm_{11}^T\adj{\ww}{0}+\mm_{21}^T\adj{\vv}{0} \\
\mm_{12}^T\adj{\ww}{0}+\mm_{22}^T\adj{\vv}{0}
\end{array}\right]
\end{equation}\label{eq:adj-sd-mul-ori}

Spatial dyad: mul-inv():
\begin{equation}
\left[
\begin{array}{c}
\ww_0\\
\vv_0
\end{array}\right]=
sd_{mul-inv}(
\left[
\begin{array}{cc}
\mm_{11} & \mm_{12}\\
\mm_{21} & \mm_{22}
\end{array}\right],
\left[
\begin{array}{c}
\ww\\
\vv
\end{array}\right])=
\left[
\begin{array}{c}
\mm_{11}^T\ww+\mm_{12}^T\vv\\
\mm_{21}^T\ww+\mm_{22}^T\vv
\end{array}\right]
\end{equation}\label{ed:sd-mul-inv}

\begin{equation}
\left[
\begin{array}{cc}
\adj{\mm}{11}&\adj{\mm}{12}\\
\adj{\mm}{21}&\adj{\mm}{22}
\end{array}\right]=
\left[
\begin{array}{cc}
\ww\adj{\ww}{0}^T & \vv\adj{\ww}{0}^T \\
\ww\adj{\vv}{0}^T & \vv\adj{\vv}{0}^T
\end{array}\right]
,
\left[
\begin{array}{c}
\adj{\ww}{}\\
\adj{\vv}{}
\end{array}\right]=
\left[
\begin{array}{c}
\mm_{11}\adj{\ww}{0}+\mm_{21}\adj{\vv}{0} \\
\mm_{12}\adj{\ww}{0}+\mm_{22}\adj{\vv}{0}
\end{array}\right]
\end{equation}\label{eq:adj-sd-mul-inv}

Spatial dyad: $\vv\vv^T()$:
\begin{equation}
\left[
\begin{array}{cc}
\mm_{11} & \mm_{12}\\
\mm_{21} & \mm_{22}
\end{array}\right]=
sd_{vvT}(
\left[
\begin{array}{c}
\ww_1 \\
\vv_1
\end{array}\right],
\left[
\begin{array}{c}
\ww_2\\
\vv_2
\end{array}\right])=
\left[
\begin{array}{cc}
\ww_1\ww_2^T & \ww_1\vv_2^T\\
\vv_1\ww_2^T & \vv_1\vv_2^T
\end{array}\right]
\end{equation}\label{ed:sd-vvt}

\begin{equation}
\left[
\begin{array}{c}
\adj{\ww}{1}\\
\adj{\vv}{1}
\end{array}\right]=
\left[
\begin{array}{cc}
\adj{\mm}{11}\ww_2+\adj{\mm}{12}\vv_2 \\
\adj{\mm}{21}\ww_2+\adj{\mm}{22}\vv_2
\end{array}\right]
,
\left[
\begin{array}{c}
\adj{\ww}{2}\\
\adj{\vv}{2}
\end{array}\right]=
\left[
\begin{array}{cc}
\adj{\mm}{11}^T\ww_1+\adj{\mm}{21}^T\vv_1 \\
\adj{\mm}{21}^T\ww_1+\adj{\mm}{22}^T\vv_1 
\end{array}\right]
\end{equation}\label{eq:adj-sd-vvt}

Spatial dyad: $\vv_\times$:
\begin{equation}
\left[
\begin{array}{ccc}
\mm_{00} & \mm_{01} & \mm_{02}\\
\mm_{10} & \mm_{11}& \mm_{12} \\
\mm_{20} & \mm_{21} & \mm_{22}
\end{array}\right]
= ([\vv_{0}, \vv_{1}, \vv_{2}]^T)_\times
=
\left[
\begin{array}{ccc}
0 & -\vv_{2} & \vv_{1}\\
\vv_{2} & 0& -\vv_{0} \\
-\vv_{1} & \vv_{0} & 0
\end{array}\right]
\end{equation}

\begin{equation}
\left[
\begin{array}{c}
\adj{\vv}{0}\\
\adj{\vv}{1}\\
\adj{\vv}{2}
\end{array}\right]
=
\left[
\begin{array}{c}
-\adj{\mm}{12}+\adj{\mm}{21}\\
\adj{\mm}{02}-\adj{\mm}{20} \\
-\adj{\mm}{01} +\adj{\mm}{10}
\end{array}\right]
\end{equation}

Spatial dyad: st2sd():
\begin{equation}
\left[
\begin{array}{cc}
\nn_{11} & \nn_{12}\\
\nn_{21} & \nn_{22}
\end{array}\right]
=
sd_{st2sd}(
\left[
\begin{array}{c}
\EE\\
\rr
\end{array}\right])=
\left[
\begin{array}{cc}
\EE & \mathbf{0}\\
-\EE\rr_\times & \EE
\end{array}\right]
\end{equation}\label{ed:sd-st2sd}

\begin{equation}
\left[
\begin{array}{c}
\adj{\EE}{} \\
\adj{\rr}{}
\end{array}\right]
=
\left[
\begin{array}{c}
\adj{\nn}{11}^T+\adj{\nn}{22}^T-\rr_{\times}\adj{\nn}{21}\\
-(\EE\adj{\nn}{21})_\times
\end{array}\right]
\end{equation}\label{ed:sd-st2sd}

Spatial dyad: shift():
\begin{equation}
\nn=
sd_{shift}(
\left[
\begin{array}{cc}
\mm_{11} & \mm_{12}\\
\mm_{21} & \mm_{22}
\end{array}\right],
\left[
\begin{array}{c}
\EE\\
\rr
\end{array}\right])=
\aa^T\bb\aa(st2sd(\left[
\begin{array}{c}
\EE\\
\rr
\end{array}\right]), 
\left[
\begin{array}{cc}
\mm_{11} & \mm_{12}\\
\mm_{21} & \mm_{22}
\end{array}\right])
\label{ed:sd-shift}\end{equation}
Adjoints of shift() can be decomposed into the adjionts of $\AA^T\BB\AA()$ and $st2sd()$.

Quaternion: mul\_vec():
\begin{equation}
\left[
\begin{array}{c}
\qq_{1}.x\\
\qq_{1}.y\\
\qq_{1}.z\\
\qq_{1}.w
\end{array}\right]
=
qt_{mul\_vec}(
\left[
\begin{array}{c}
\qq_{2}.x\\
\qq_{2}.y\\
\qq_{2}.z\\
\qq_{2}.w
\end{array}\right],
\left[
\begin{array}{c}
\vv.x\\
\vv.y\\
\vv.z
\end{array}\right])=
\left[
\begin{array}{c}
\qq_{2}.w * \vv.x + \qq_{2}.y * \vv.z - \qq_{2}.z * \vv.y\\
\qq_{2}.w * \vv.y + \qq_{2}.z * \vv.x - \qq_{2}.x * \vv.z\\
\qq_{2}.w * \vv.z + \qq_{2}.x * \vv.y - \qq_{2}.y * \vv.x\\
-\qq_{2}.x * \vv.x - \qq_{2}.y * \vv.y - \qq_{2}.z * \vv.z
\end{array}\right]
\end{equation}\label{ed:qt-mul-vec}

\begin{equation}
\left[
\begin{array}{c}
\adj{\qq}{2}.x\\
\adj{\qq}{2}.y\\
\adj{\qq}{2}.z\\
\adj{\qq}{2}.w
\end{array}\right]
=
\left[
\begin{array}{c}
\adj{\qq}{1}.x*q.w - \adj{\qq}{1}.y*q.z + \adj{\qq}{1}.z*q.y - \adj{\qq}{1}.w*q.x\\
\adj{\qq}{1}.x*q.z + \adj{\qq}{1}.y*q.w - \adj{\qq}{1}.z*q.x - \adj{\qq}{1}.w*q.y\\
-\adj{\qq}{1}.x*q.y + \adj{\qq}{1}.y*q.x + \adj{\qq}{1}.z*q.w - \adj{\qq}{1}.w*q.z
\end{array}\right]
\end{equation}\label{ed:qt-mul-vec-adj}

\begin{equation}
\left[
\begin{array}{c}
\vv.x\\
\vv.y\\
\vv.z
\end{array}\right]=
\left[
\begin{array}{c}
\adj{\qq}{1}.x*\qq_2.w - \adj{\qq}{1}.y*\qq_2.z + \adj{\qq}{1}.z*\qq_2.y - \adj{\qq}{1}.w*\qq_2.x\\
\adj{\qq}{1}.x*\qq_2.z + \adj{\qq}{1}.y*\qq_2.w - \adj{\qq}{1}.z*\qq_2.x - \adj{\qq}{1}.w*\qq_2.y\\
-\adj{\qq}{1}.x*\qq_2.y + \adj{\qq}{1}.y*\qq_2.x + \adj{\qq}{1}.z*\qq_2.w - \adj{\qq}{1}.w*\qq_2.z
\end{array}\right]
\end{equation}\label{ed:qt-mul-vec-adj2}

Quaternion: mul\_qt():
\begin{equation}
\left[
\begin{array}{c}
\qq_{}.x\\
\qq_{}.y\\
\qq_{}.z\\
\qq_{}.w
\end{array}\right]
=
qt_{mul\_qt}(
\left[
\begin{array}{c}
\qq_{1}.x\\
\qq_{1}.y\\
\qq_{1}.z\\
\qq_{1}.w
\end{array}\right],
\left[
\begin{array}{c}
\qq_{2}.x\\
\qq_{2}.y\\
\qq_{2}.z\\
\qq_{2}.w
\end{array}\right])=
\left[
\begin{array}{c}
\qq_1.w * \qq_2.x + \qq_1.x * \qq_2.w + \qq_1.y * \qq_2.z -\qq_1.z * \qq_2.y\\
\qq_1.w * \qq_2.y + \qq_1.y * \qq_2.w + \qq_1.z * \qq_2.x -\qq_1.x * \qq_2.z\\
\qq_1.w * \qq_2.z + \qq_1.z * \qq_2.w + \qq_1.x * \qq_2.y -\qq_1.y * \qq_2.x\\
\qq_1.w * \qq_2.w - \qq_1.x * \qq_2.x - \qq_1.y * \qq_2.y -\qq_1.z * \qq_2.z
\end{array}\right]
\end{equation}\label{ed:qt-mul-qt}

\begin{equation}
\left[
\begin{array}{c}
\adj{\qq}{1}.x\\
\adj{\qq}{1}.y\\
\adj{\qq}{1}.z\\
\adj{\qq}{1}.w
\end{array}\right]
=
\left[
\begin{array}{c}
\adj{\qq}{}.x*\qq_2.w - \adj{\qq}{}.y*\qq_2.z + \adj{\qq}{}.z*\qq_2.y - \adj{\qq}{}.w*\qq_2.x\\
\adj{\qq}{}.x*\qq_2.z + \adj{\qq}{}.y*\qq_2.w - \adj{\qq}{}.z*\qq_2.x - \adj{\qq}{}.w*\qq_2.y\\
-\adj{\qq}{}.x*\qq_2.y+ \adj{\qq}{}.y*\qq_2.x + \adj{\qq}{}.z*\qq_2.w - \adj{\qq}{}.w*\qq_2.z\\
\adj{\qq}{}.x*\qq_2.x + \adj{\qq}{}.y*\qq_2.y + \adj{\qq}{}.z*\qq_2.z + \adj{\qq}{}.w*\qq_2.w
\end{array}\right]
\end{equation}\label{ed:qt-mul-qt-adj}

\begin{equation}
\left[
\begin{array}{c}
\adj{\qq}{2}.x\\
\adj{\qq}{2}.y\\
\adj{\qq}{2}.z\\
\adj{\qq}{2}.w
\end{array}\right]
=
\left[
\begin{array}{c}
\adj{\qq}{}.x*\qq_1.w + \adj{\qq}{}.y*\qq_1.z - \adj{\qq}{}.z*\qq_1.y - \adj{\qq}{}.w*\qq_1.x\\
-\adj{\qq}{}.x*\qq_1.z+ \adj{\qq}{}.y*\qq_1.w + \adj{\qq}{}.z*\qq_1.x - \adj{\qq}{}.w*\qq_1.y\\
\adj{\qq}{}.x*\qq_1.y - \adj{\qq}{}.y*\qq_1.x + \adj{\qq}{}.z*\qq_1.w - \adj{\qq}{}.w*\qq_1.z\\
\adj{\qq}{}.x*\qq_1.x + \adj{\qq}{}.y*\qq_1.y + \adj{\qq}{}.z*\qq_1.z + \adj{\qq}{}.w*\qq_1.w
\end{array}\right]
\end{equation}\label{ed:qt-mul-qt-adj2}

Please refer to our code for more details.

\end{appendices}

\end{document}